\documentclass[10pt,notitlepage]{article}
\usepackage{epsfig}
\usepackage{amssymb}

\newcommand{\be}{\begin{equation}}
\newcommand{\ee}{\end{equation}}
\newcommand{\bea}{\begin{eqnarray}}
\newcommand{\eea}{\end{eqnarray}}

\usepackage{bm}
\usepackage{color}
\usepackage{amsmath}
\usepackage{graphicx}
\usepackage{amssymb}
\usepackage{cite}

\newcommand{\red}[1]{{\color{red}#1}}

\begin{document}

\title{It's a super deal -- \\ train recurrent network on {\it noisy} data \\
and get {\it smooth} prediction free}
\author{Boris  Rubinstein,
\\Stowers Institute for Medical Research
\\1000 50$^{}\mbox{th}$ St., Kansas City, MO 64110, U.S.A.}
\date{\today}
\maketitle

\begin{abstract}
Recent researches demonstrate that
prediction of time series by predictive recurrent neural networks based on the 
noisy input generates a {\it smooth} anticipated trajectory. We examine 
influence of the noise component in both the training data sets and 
the input sequences on network prediction quality. We propose and discuss
an explanation of the observed noise compression in the predictive
process. We also discuss importance of this property of recurrent networks in the 
neuroscience context for the evolution of living organisms.
\end{abstract}


\section{Introduction}

Recurrent neural networks (RNNs)  due to
their ability to process ordered sequences of data 
have found applications in many fields 
of science, engineering and humanities.
RNNs represent the most important elements of 
predictive recurrent networks (PRN).
One of the PRN applications is time series prediction used
in analysis of business and financial data, weather forecast {\it etc}.
It is believed that animal brains also use some kind of PRNs to predict
moving object trajectories.
Trajectory prediction based on incomplete 
or noisy data is one of the 
most amazing features of organism brains that allows living creatures to
survive in complex and usually unfriendly environment.

What does happen when a smooth trajectory is perturbed by an
external noise of specific statistics, {\it e.g.}, white noise? How would PRN
extrapolate the input of such noisy time series?
Generally speaking, when a noisy signal is used as an input to a PRN
it is expected that a trained network would be able to extrapolate the {\it noisy}
time series. 
It appears that the extrapolated trajectory is not noisy --
filtering of the noisy perturbation of the Lorenz attractor dynamics is reported 
in recurrent multi-layer perception network --
the reconstructed signals are "reasonably close to the noise-free
signal and the iterated predictions are smoother in comparison to the noisy signals" 
\cite{Haykin2001}.
This observation leads to the following question -- given a smooth deterministic 
function with added noise component as a PRN input will the 
trajectory anticipated by the network be noisy or smooth?
A short note \cite{Yeo2019} considers LSTM network \cite{Hochreiter1997} 
with small number (128) of neurons trained on the 
Mackey-Glass time series with added noise. It appears that 
with the increase of the noise level PRN behavior depends more on its own 
dynamics than on the input data. On the contrary, the training using the 
noiseless input produces PRN with very high sensitivity to small perturbations.

The author of this manuscript reported earlier that PRN trained
on segments of \emph{noisy} trajectory and being fed a segment of such trajectory
produces a \emph{smooth} extrapolating curve and this effect is
independent of the PRN architecture, size and depth \cite{Rub2020a,Rub2020b}. It was shown that
PRNs do not {\it filter out} the noise component of the input
sequence but somehow predicts a smooth curve that is quite close
to the actual noise-free dynamics.
The effect is observed for different training algorithms and prediction procedures 
\cite{Rub2020a,Rub2020b,Rub2021a}.
In this manuscript we investigate an influence of the 
noise amplitude in the training data sets and input sequences
on the network prediction of actual trajectory. 
We also discuss a possible reason for the observed PRN ability
to generate a smooth time series and argue that it
is related to the PRN training procedure and enhanced by 
the predictive algorithms. 
In this manuscript the term {\it "smooth"} for
description of the predicted trajectories means that
{\it the deviation from the actual trajectory is 
negligible compared to the noise amplitude of the 
input sequence fed into the network}.

\section{Recurrent network training and predictive algorithms}
\label{algorithm0}

A typical PRN is designed to use an input sequence $\bm X = \{\bm x_i\}, \ 1\le i \le m$ of 
$m$ elements $\bm x_i$ that represents 
a segment of the time series in order to 
generate a predicted sequence of $p$ elements $\bm x_{m+j},\ 1\le j \le p$
determining the consecutive part of the same time series,
so that the elements of both the 
input and predicted sequences should have similar structure 
(numerical vectors, words, symbols, images, musical notes {\it etc.}).

\subsection{Network training}

To perform this task PRN should be first trained using a training set consisting of
sequences of fixed or variable length $m$, 
each sequence $\bm X = \{\bm x_i\}$ is accompanied by either a 
single next element $\bm x_{m+1}$ for "seq-to-one" PRNs \cite{Rub2020a,Rub2020b} or 
a sequence of elements $\{\bm x_i\}, \ m+1\le i \le m+l$
for "seq-to-seq" network \cite{Rub2021a}. In this manuscript we focus 
on "seq-to-one" RNNs.
In this case for each training set  $\bm X^k$ PRN produces a prediction  $\bar{\bm x}_{m+1}^k$ that
compared to the actual values ${\bm x}_{m+1}^k$
The network parameters
are fitted to minimize the mean square difference (a training error $E_{tr}$) after $N$ training rounds
\be
E_{tr}=(1/N)\sum_{k=1}^N |\bar{\bm x}_{m+1}^k-{\bm x}_{m+1}^k|^2.
\nonumber 
\ee
Once the network is trained it can be used to generate several
consecutive values  $\bar{\bm x}_{m+j}, \ 1 \le j \le p$.

The mode of prediction for "seq-to-one" PRNs is the following.
An inner state (an activation level) of the network $n$ neurons is described by a
$n$-dimensional vector $\bm s$. Its discrete dynamics 
is usually governed by a vector map
\be
\bm s_i = \bm F(\bm x_i, \bm s_{i-1}, \bm P),
\label{map}
\ee
where $\bm F$ determines the network architecture and
$\bm P$ embraces the network trainable parameters. Assuming these
parameters fixed for the trained network we drop this argument below.
This algorithm describes the neurons having no memory
so that the current state $\bm s_i$ of the network depends only
on the current input signal $\bm x_i$ and the previous RNN state 
$\bm s_{i-1}$ while the parameter values are fixed during the 
prediction process. The last state $\bm s_m$ of the network is transformed linearly
to predict a single point $\bar{\bm x}_{m+1}$
\be
\bar{\bm x}_{m+1} = \bm L(\bm s_m) = \bm W \cdot \bm s_m + \bm b,
\label{predlin}
\ee
where $\bm W$ is a matrix and $\bm b$ is a bias vector
both have trainable parameters.

\subsection{Prediction algorithms}

When the network is trained on the sequences of the fixed length the standard
predictive algorithm uses a "moving window" (MW) recursion. 
One starts with a sequence $\bm X^{1}$ of length $m$ supplied as an input
to the network; it leads to generation of a predicted point 
$\bar{\bm x}_{m+1} $. The next input sequence $\bm X^{2}$ 
is produced by dropping the first point of $\bm X^{1}$ and 
adding the predicted point $\bar{\bm x}_{m+1}$ to the result.
This sequence is used as a new input leading to generation of $\bar{\bm x}_{m+2}$
and the next input ${\bm X}^{3}$ is formed.
Thus at $k$-th predictive step the input $\bm X^k$ to PRN is 
formed by adding to the original input $\bm X^1$ all previously $k-1$
predicted points and shifting the window by $k-1$ steps forward.
The recursive procedure is repeated $p$ times to produce a sequence of  $p$ points
$\bar{\bm x}_{m+j}, \ 1 \le j \le p$ approximating the sequence
$\{\bm x_i\}$ for $m+1 \le i \le m+p$. 
It should be noted that this approach is used to produce all the figures in the next section.

For network training with sequences of the variable length 
(not larger than some $M$) the 
MW predictive algorithm can be modified into the "expanding window"
(EW) version. After each predictive round the newly generated
point is added to the original input sequence, so that after the $k$-th prediction 
round the length of the input sequence $\bm X^k$ 
is $m+k$, where $m$ denotes the size of the 
initial input sequence and $m+k \le M$. The main reason of EW algorithm application is that
a gradual increase of the input length usually
leads to better prediction quality and it also produces
smoother anticipated trajectories.

It should be underlined that both above algorithms require 
some memory unit to store the input sequence values and thus they 
perform their task {\it non-autonomously}. From the 
biological perspective the presence of such a unit requires
additional resources that might not be available. The author recently
suggested \cite{Rub2020a,Rub2020b} a memoryless (ML) algorithm generating 
smooth trajectories that coincide with high 
accuracy with the trajectories predicted by the traditional algorithms.
This {\it autonomous} algorithm can be applied for any (even untrained) predictive network.
In Appendix we show that ML algorithm is just a "compression" 
of the regular EW procedure.

\subsection{Training set and prediction input sequences}
As PRNs process discrete data sets (time series) and we are interested in 
the noise influence on the network prediction quality we have to 
define sequences that describe smooth trajectories with
(significant) noise components added. 
Introduce a smooth continuous vector function $\bm f(t)$ of the 
scalar argument representing an actual trajectory in the phase space.
An apparent trajectory fed into a network is defined by
$\bm g(t,a) = \bm f(t) + a \bm \xi (t)$ with the noise component 
$\bm \xi (t)$ of specific statistics (say, white noise) and amplitude $a$. 
We assume that the noise average vanishes $\langle \bm \xi (t) \rangle=0$.
Define a fixed or variable time step $\Delta t$ and generate 
a sequence $\bm G(a) = \{\bm g_j(a)\}$ where 
$\bm g_j(a) =\bm g(t_0+ j \Delta t,a) = \bm f_j + a \bm \xi_j$.
The training sets are constructed from the sequence $\bm G(a_0)$
and the input sets used for the prediction by a trained PRN 
represent the sequence $\bm G(a_i)$  where 
$0 \le a_0,a_i \le a_m$. The maximal noise amplitude $a_m$ 
assumed to be comparable to the characteristic range of values of the 
smooth function $\bm f(t)$.

\section{RNN training and performance}
\label{training}

The network training with noisy input sequences requires multiple 
data sets that have the same actual smooth trajectory $\bm f(t)$ or
sequence $\bm f_j$ 
and 
perturbations $a_0 \bm \xi (t)$ or $a_0 \bm \xi_j$.

\subsection{Network architecture and training data sets}
The PRNs with a single recurrent basic or 
LSTM layer \cite{Hochreiter1997} have a small number $n=20$ of neurons.
For each trajectory and given noise amplitude $a_0=0, 0.15, 0.4, 0.75$ 
the training set is constructed by generating $6000$ segments of 
variable length ($5 \le m \le 50$). 

We considered two qualitatively different types of trajectories.
In order to save computational resources one can use a periodic function 
$\bm f(t)$ with hundreds of periods to generate a discrete sequence 
$\bm f_j = \bm f(t_j)$ and add to each point of this sequence 
$a_0 \bm \xi_j$ where $\bm \xi_j = \bm \xi(t_j)$ where the amplitudes 
of the individual components of the vector $\bm \xi$ might differ. 
Thus, network of the first type is trained on two smooth periodic one-dimensional functions
-- the sine wave $\bm f(t) = f(t) = \sin(2\pi t)$ and 
the shifted triangle wave
$\bm f(t) = f(t)  = 1/2 + 1/\pi \arcsin (\sin 2\pi t)$ with added noise
$a_0 \xi (t)$.
The time step $\Delta t$ between the adjacent time points is selected equal to
$\Delta t = 0.01$. 
The training procedure is performed for 50 epochs on the two merged sets 
with fixed noise amplitude $a_0$ having $12000$
segments with $20\%$ validation set using Adam algorithm. 
This way we produce individual network for each value of $a_0$.

We also consider parabolic trajectories that with high accuracy 
resemble typical trajectories in nature describing
motion of a solid object in the gravity field like a flying rock.
Such trajectories are finite and essentially two dimensional.
Consider a parabolic trajectory with a range $b$ and vertex height $h$
described by an equation $y(x)=h (1- 4 x^2/b^2)$ with 
$y(-b/2) = y(b/2) = 0$; the vertex $\{0,h\}$ is
reached at $t=1/2$.
Assuming that points of such a trajectory are fed into a network
at equal time step $\Delta t = 0.01$ one has to use
a parametric representation of this curve 
$\bm f(t) = \{b (t-1/2), 4 h t(1-t)\}$ where $0 \le t \le 1$.
The noise component is also two dimensional 
$a_0 \bm \xi = \{a_0 \xi_x, a_0 \xi_y\}$ with $a_0=0, 0.15, 0.4, 0.75$
and $\xi_x = 0.1 \xi_y$.

A parabola can be characterized by a ratio $h/b$ of its
height $h$ to range $b$. The trajectories with 
extreme values of this ratio $h/b \ll 1$ and $h/b \gg 1$ 
present cases when accurate prediction fails.
When $h \ll b$ we have a very shallow trajectory and
superimposed noise leads to totally random and thus unpredictable 
curve. 
For $h \gg b$ the influence of noise produces
strong overlapping of ascending and descending parts of the 
curve that prevents good guessing of the impact point.
Note here that prediction also fails when a line of sight is nearly parallel
to a plane containing a parabola -- it happens mainly
due to trajectory perturbations normal to that plane.
For this reason, we use three parabolas with fixed value of vertex height $h=1$ and 
ranges $b=1,2,4$.
The training procedure is performed for 50 epochs on the three merged sets 
with fixed noise amplitude $a_0$ having $18000$
segments with $20\%$ validation set using Adam algorithm.


\subsection{Prediction results for one dimensional trajectories}
\begin{figure}[h!]
\begin{center}
\begin{tabular}{cc}
\psfig{figure=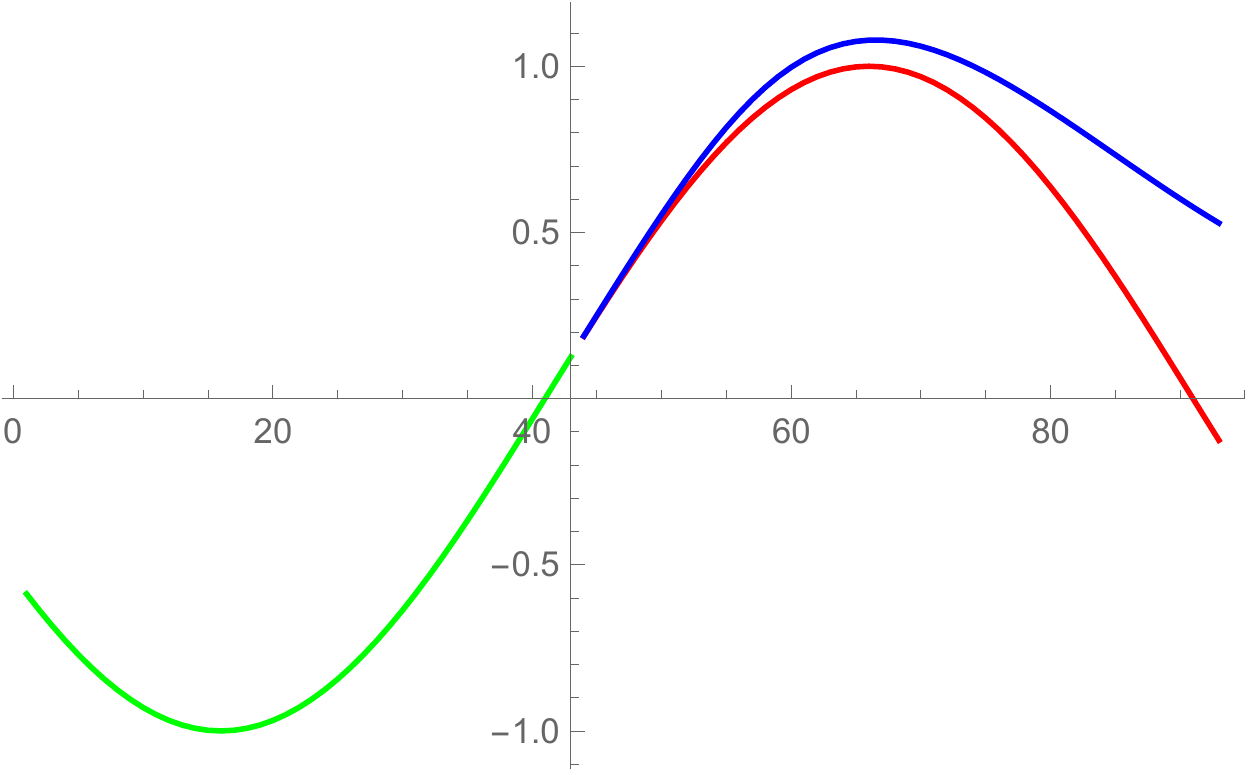,width=7cm} 
&
\psfig{figure=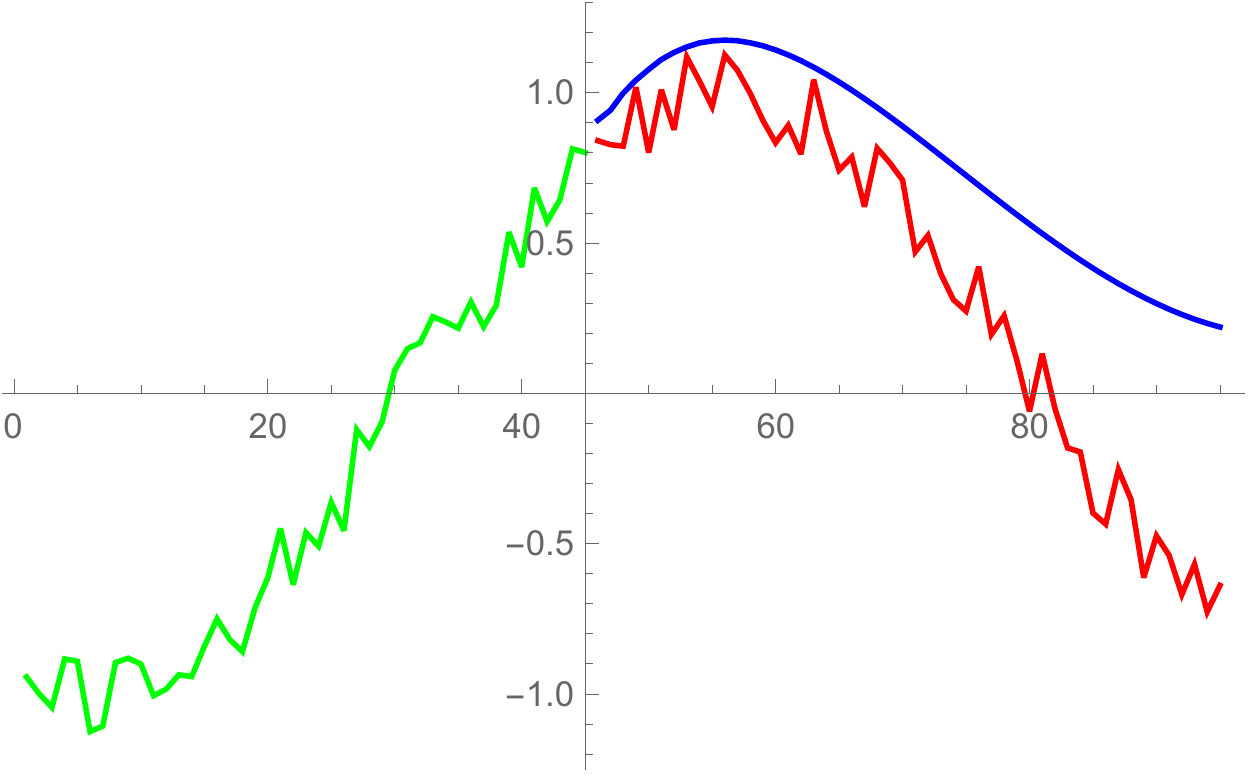,width=7cm}
\\
({\bf a}) & ({\bf b}) 
\\
\psfig{figure=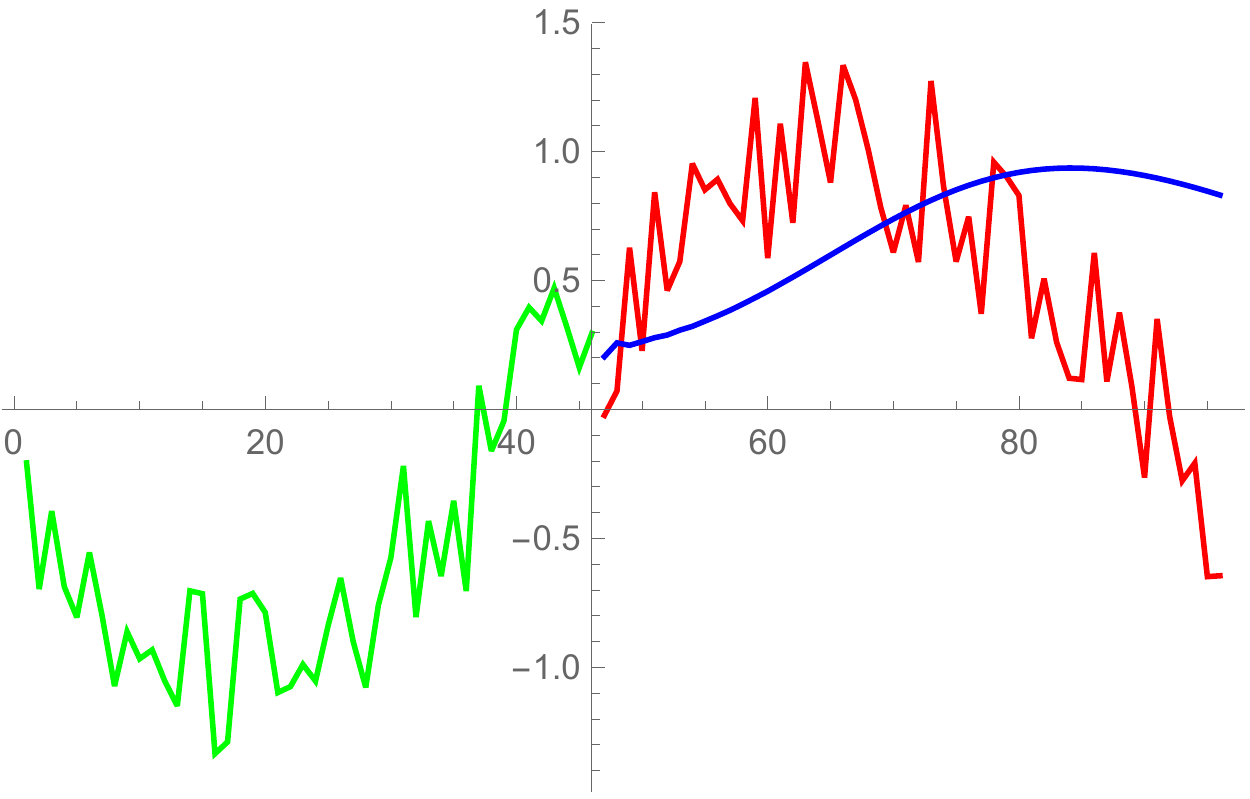,width=7cm}
&
\psfig{figure=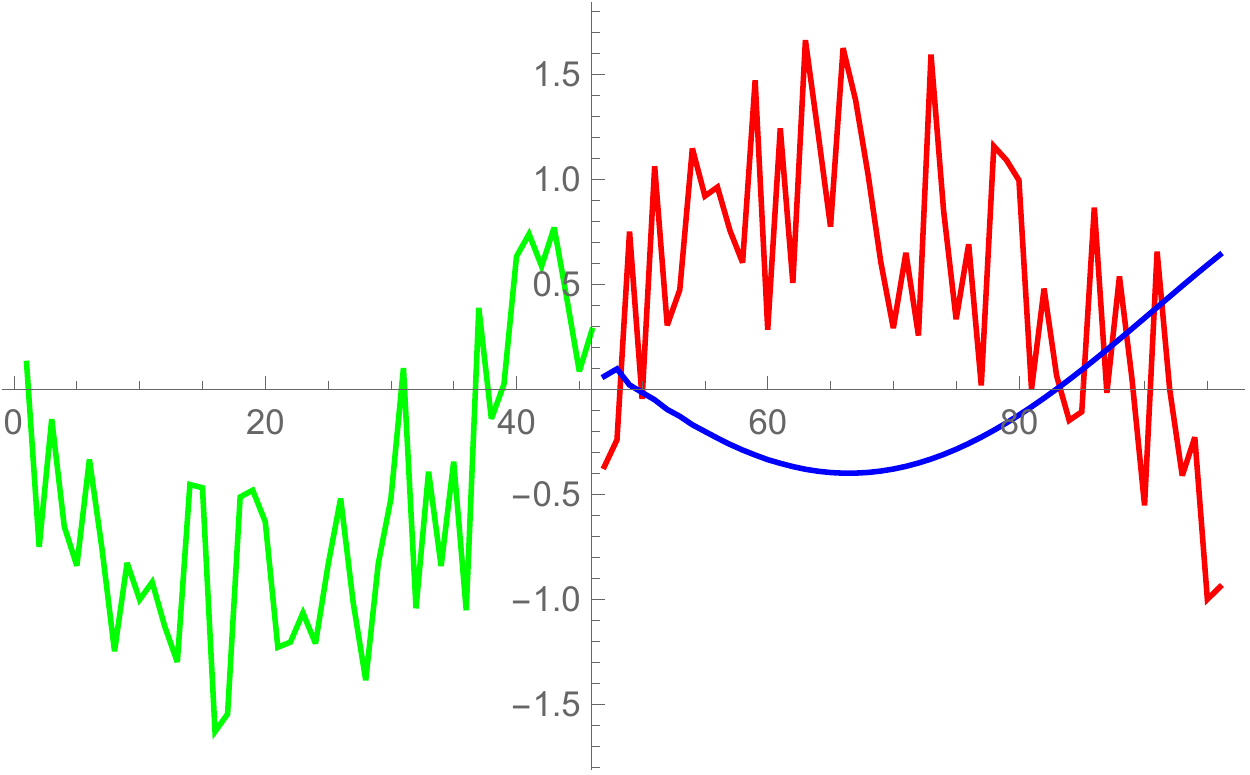,width=7cm}
\\
 ({\bf c}) & ({\bf d})
\end{tabular}
\caption{The input segment of a sequence (green) 
with ({\bf a}) $a_i=0$,  ({\bf b}) $a_i=0.15$,  ({\bf c}) $a_i=0.40$,  ({\bf d}) $a_i=0.75$ 
of sine wave, the subsequent segment of 
the data sequence (red) and predicted dynamics (blue) by basic network with $20$  neurons trained 
using noiseless data sets with $a_0=0$.
}
\label{Fig1}
\end{center}
\end{figure}

The qualitative results are the following. The PRN trained on the data sets 
based on the smooth functions $f(t)$ ($a_0=0$) predict quite good on the 
similar smooth inputs when the prediction length $p \sim m$ but for $p > m$ 
the prediction error starts to grow. They demonstrate very low prediction 
ability for inputs with the noise amplitude comparable to the 
smooth function characteristic range (Fig. \ref{Fig1}).

\begin{figure}[h!]
\begin{center}
\begin{tabular}{cccc}
\psfig{figure=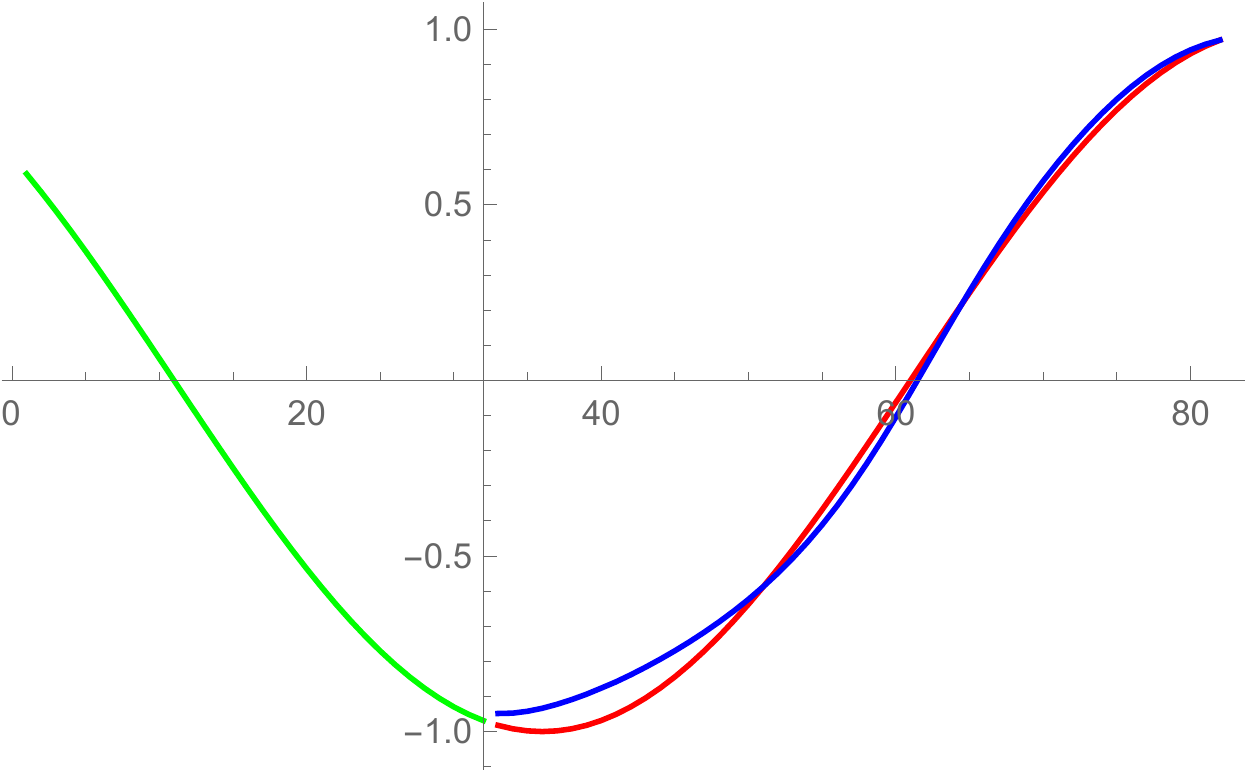,width=3.5cm} 
&
\psfig{figure=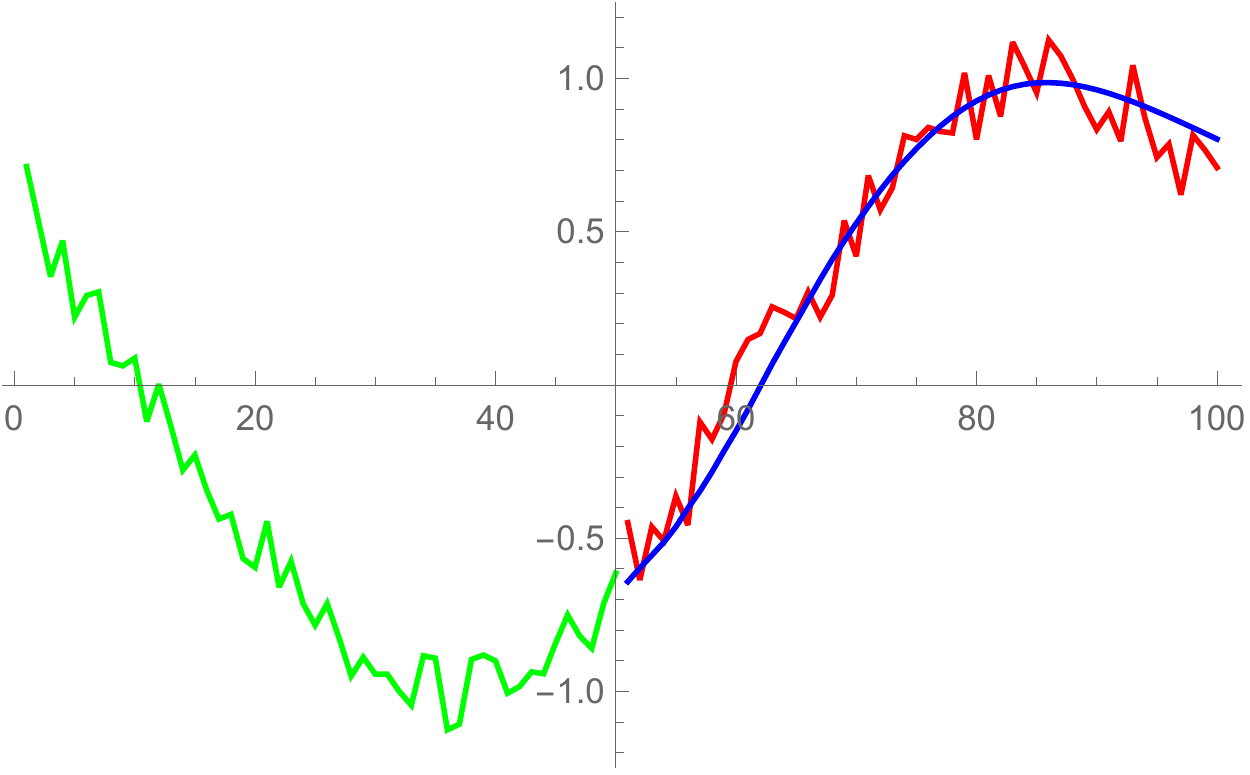,width=3.5cm}
&
\psfig{figure=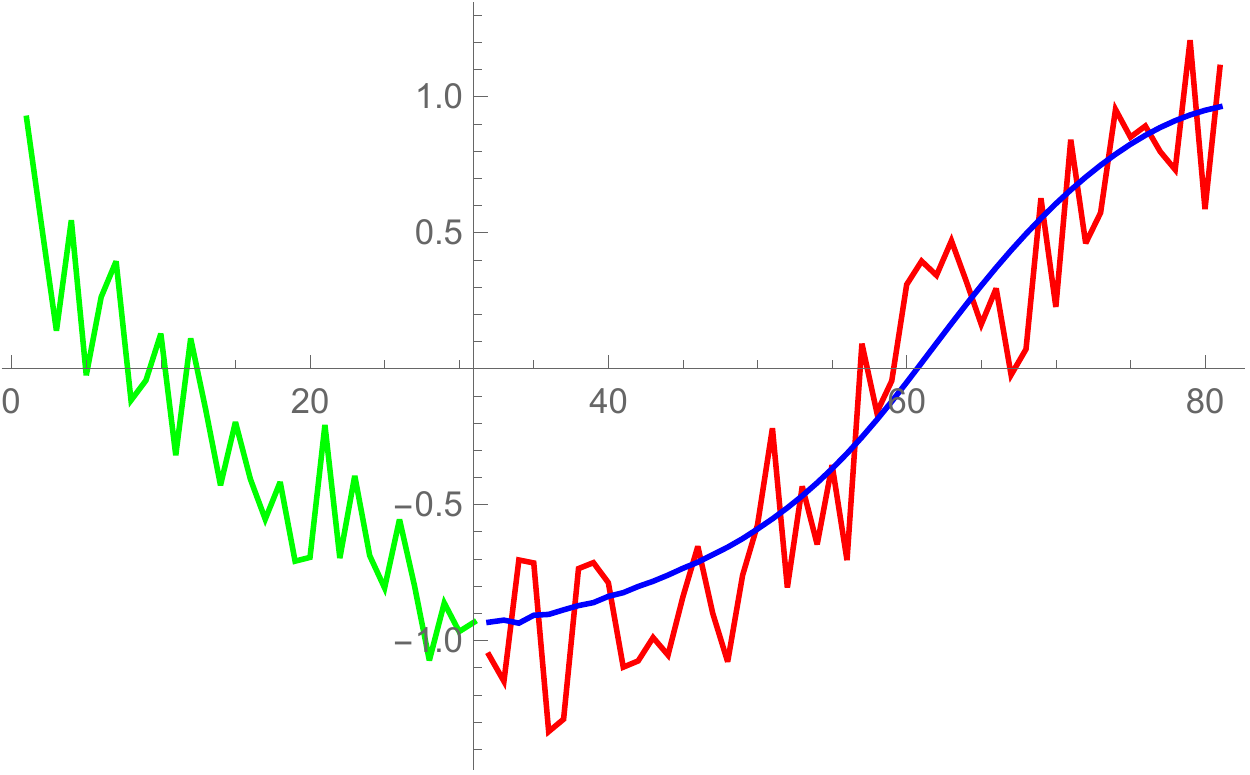,width=3.5cm}
&
\psfig{figure=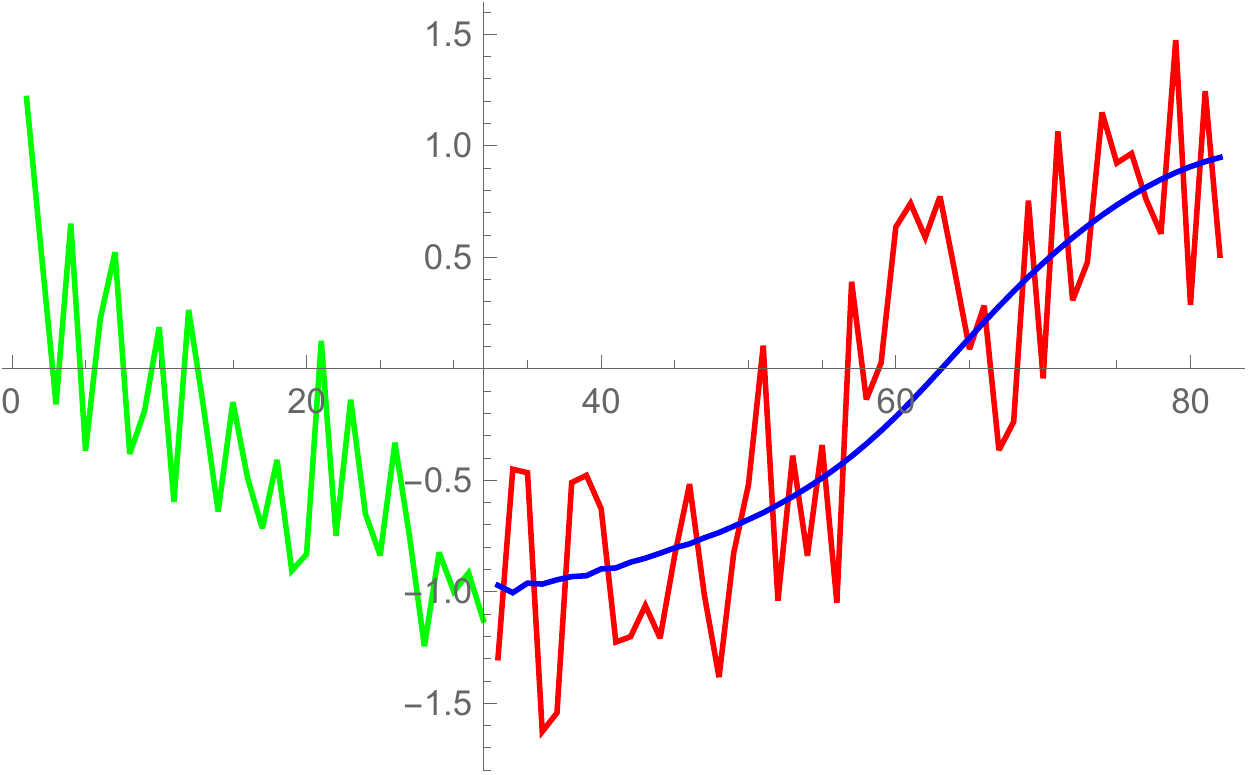,width=3.5cm}
\\
\psfig{figure=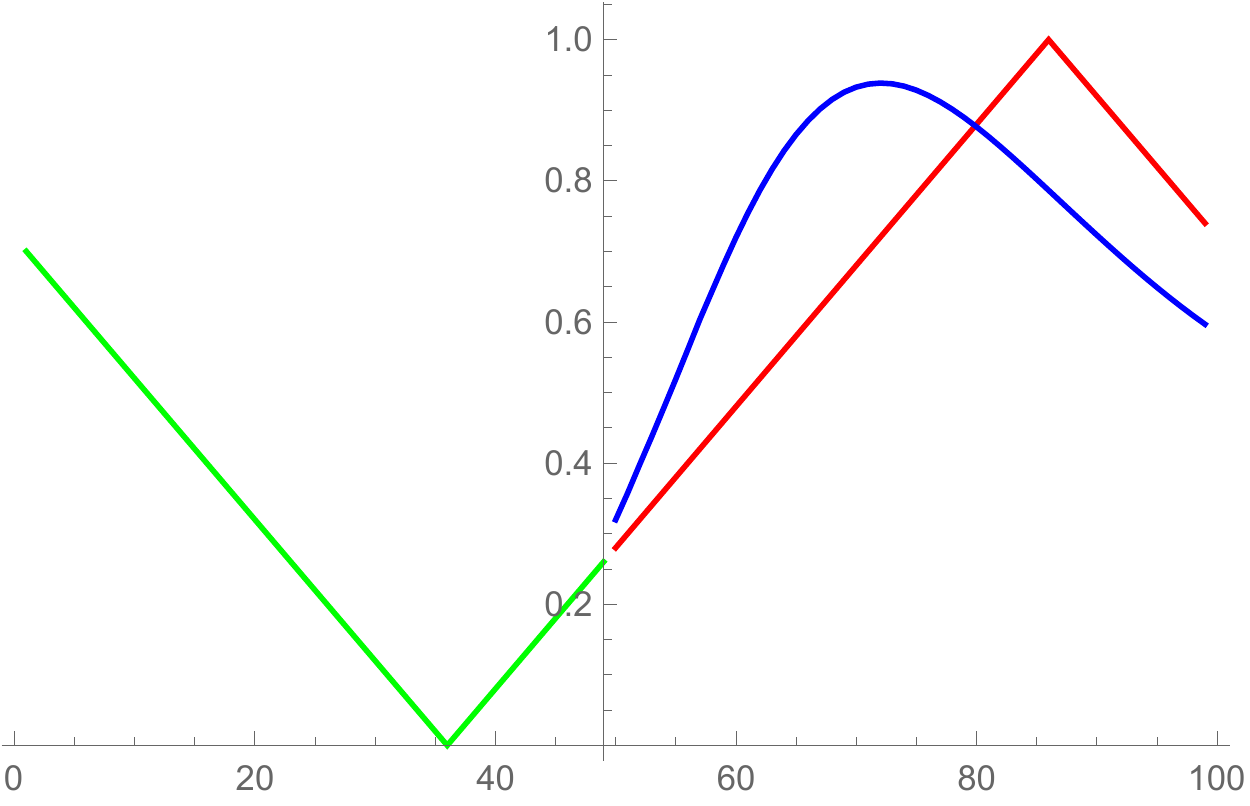,width=3.5cm} 
&
\psfig{figure=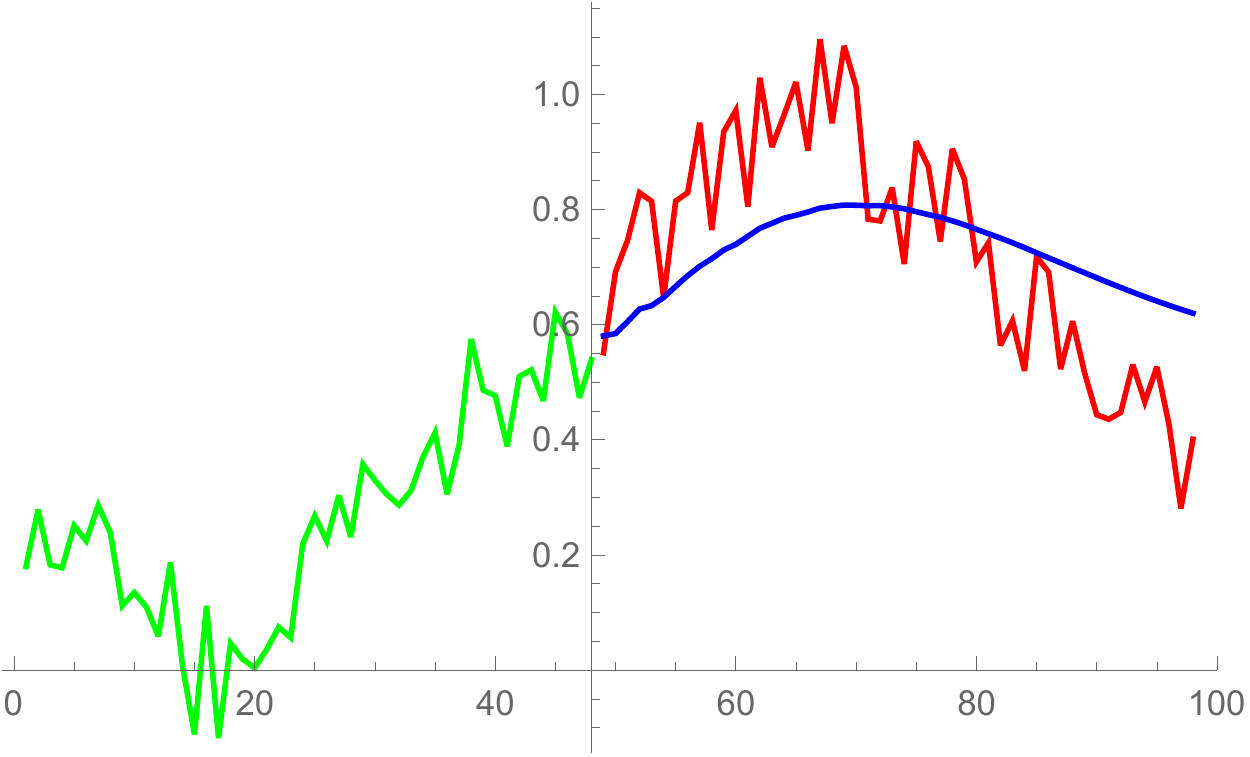,width=3.5cm}
&
\psfig{figure=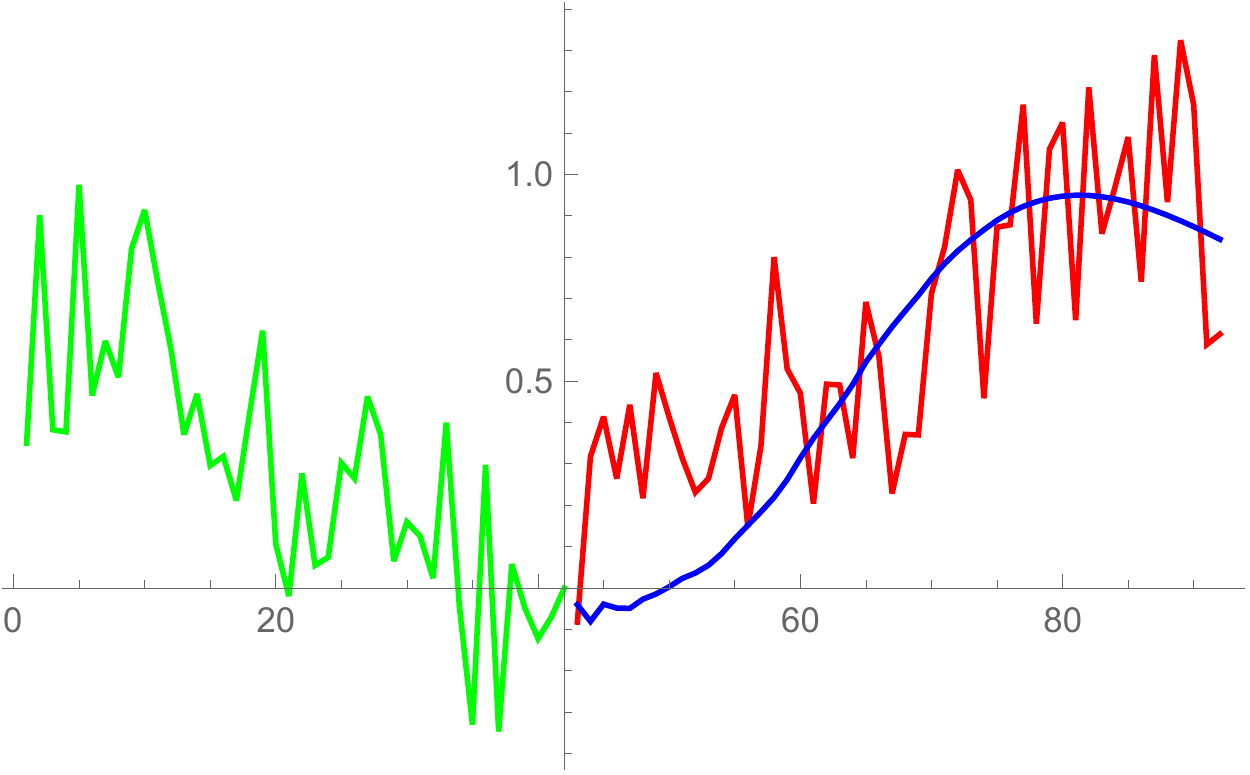,width=3.5cm}
&
\psfig{figure=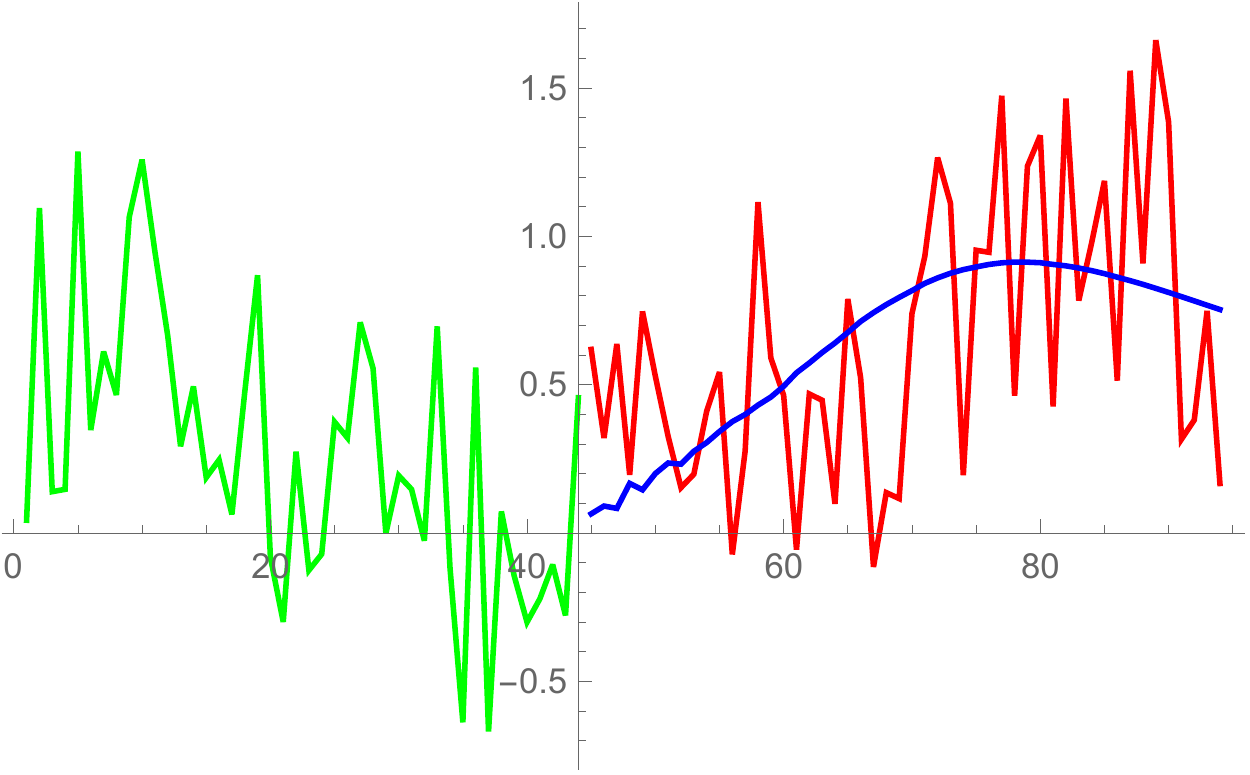,width=3.5cm}
\\
({\bf a}) & ({\bf b}) & ({\bf c}) & ({\bf d})
\end{tabular}
\caption{The input segment of a sequence (green) 
with ({\bf a}) $a_i=0$,  ({\bf b}) $a_i=0.15$,  ({\bf c}) $a_i=0.40$,  ({\bf d}) $a_i=0.75$ 
of sine (first row) and triangle (second row) wave, the subsequent segment of 
the data sequence (red)
and predicted dynamics (blue) using basic layer network with $20$  neurons trained 
using data sets with noise amplitude $a_0=0.15$.
}
\label{Fig2}
\end{center}
\end{figure}

The networks trained on the noisy data sets with $a_0 > 0$
consistently fail to predict the noisy dynamics of $\bm g(t) = g(t)$, instead
they produce some smooth predictions $\bar {\bm g}(t)=\bar g(t)$
represented by sequences $\bar {\bm g_j} = \bar g_j$ 
(few examples for $a_0=0.15$ are shown in  Fig. \ref{Fig2}).

\begin{figure}[h!]
\begin{center}
\begin{tabular}{cc}
\psfig{figure=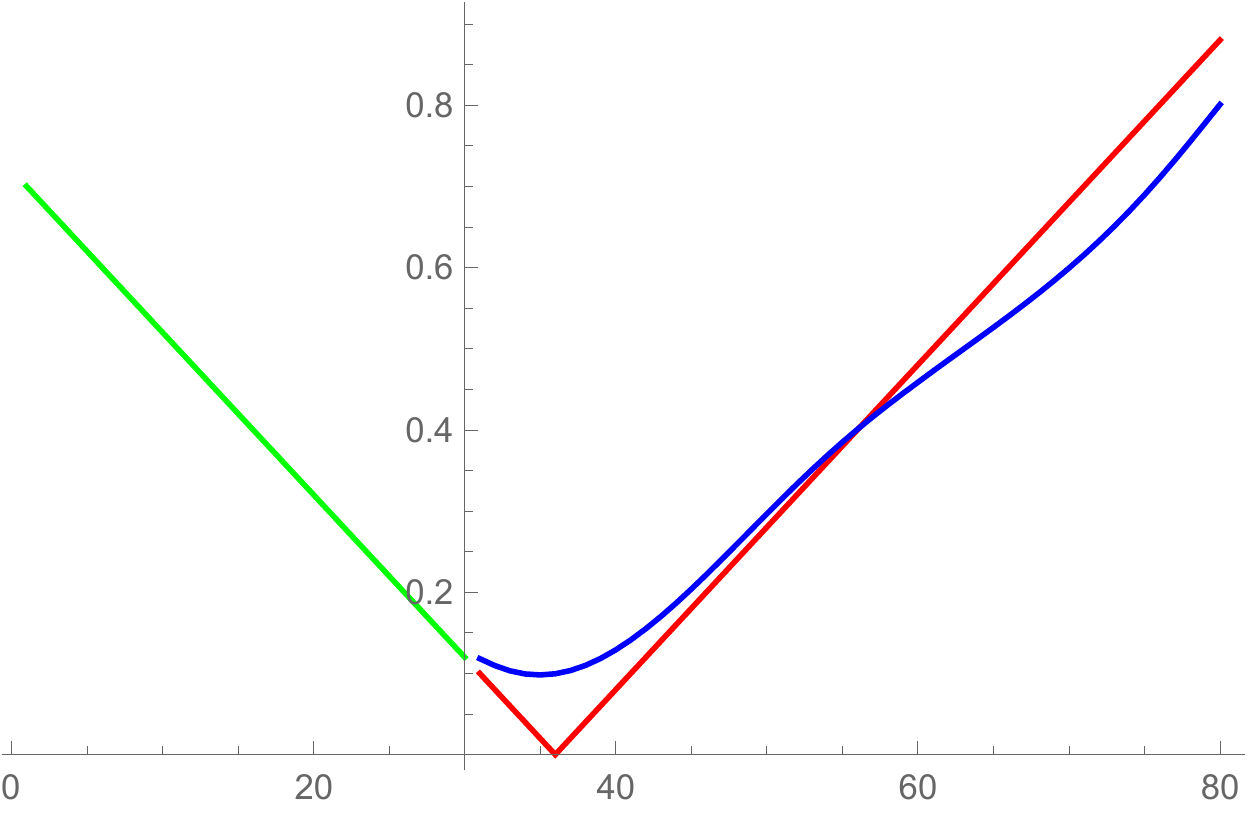,width=7cm} 
&
\psfig{figure=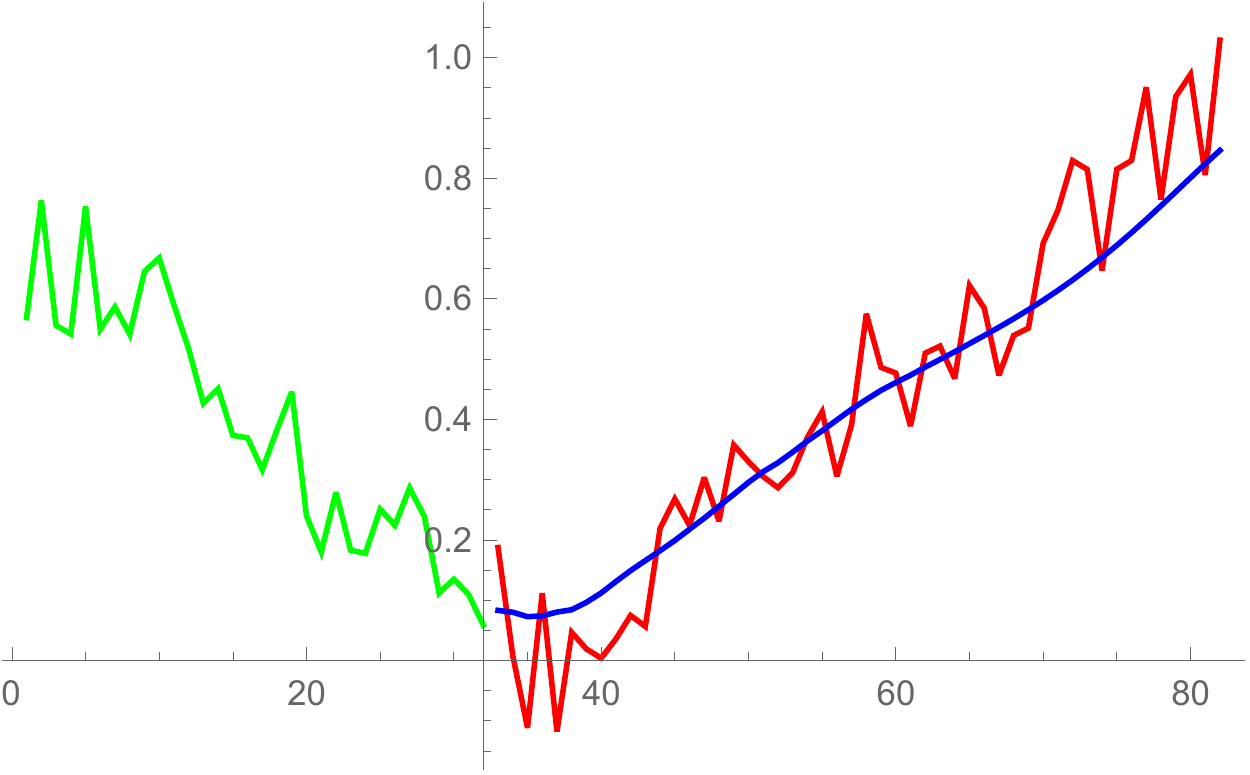,width=7cm}
\\
({\bf a}) & ({\bf b}) 
\\
\psfig{figure=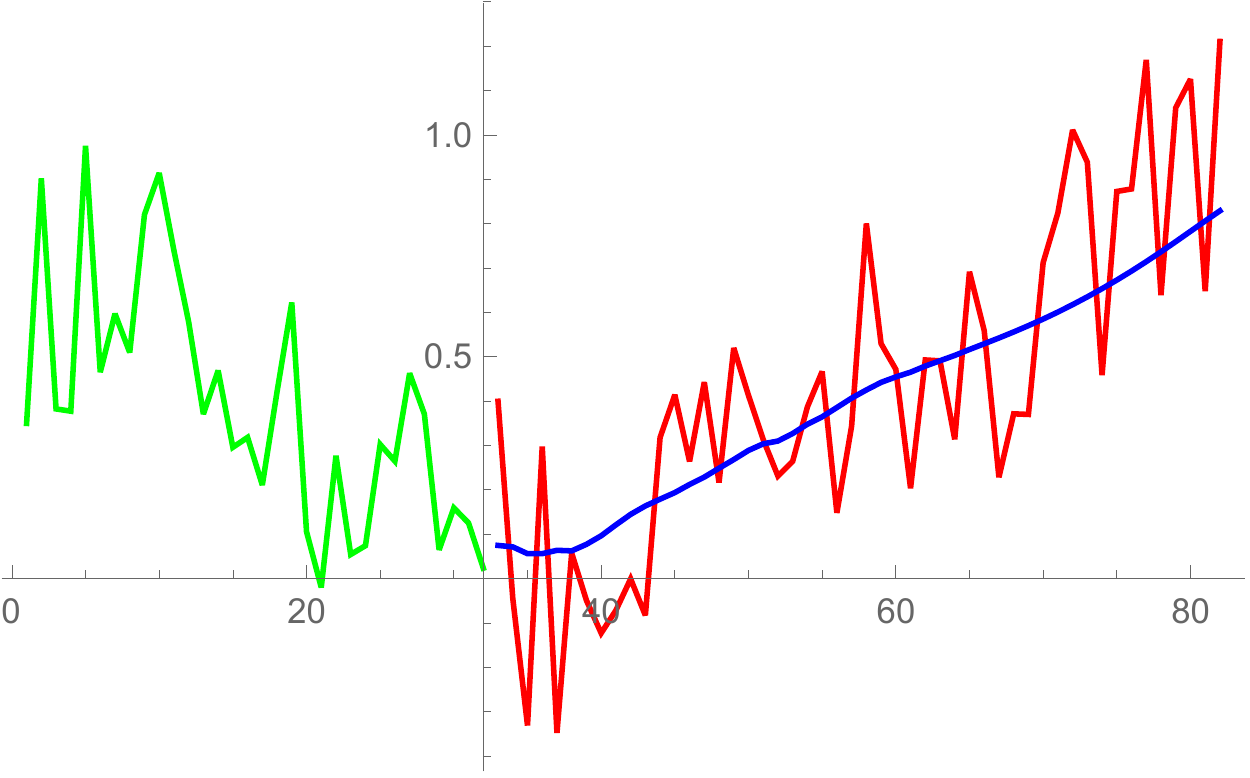,width=7cm}
&
\psfig{figure=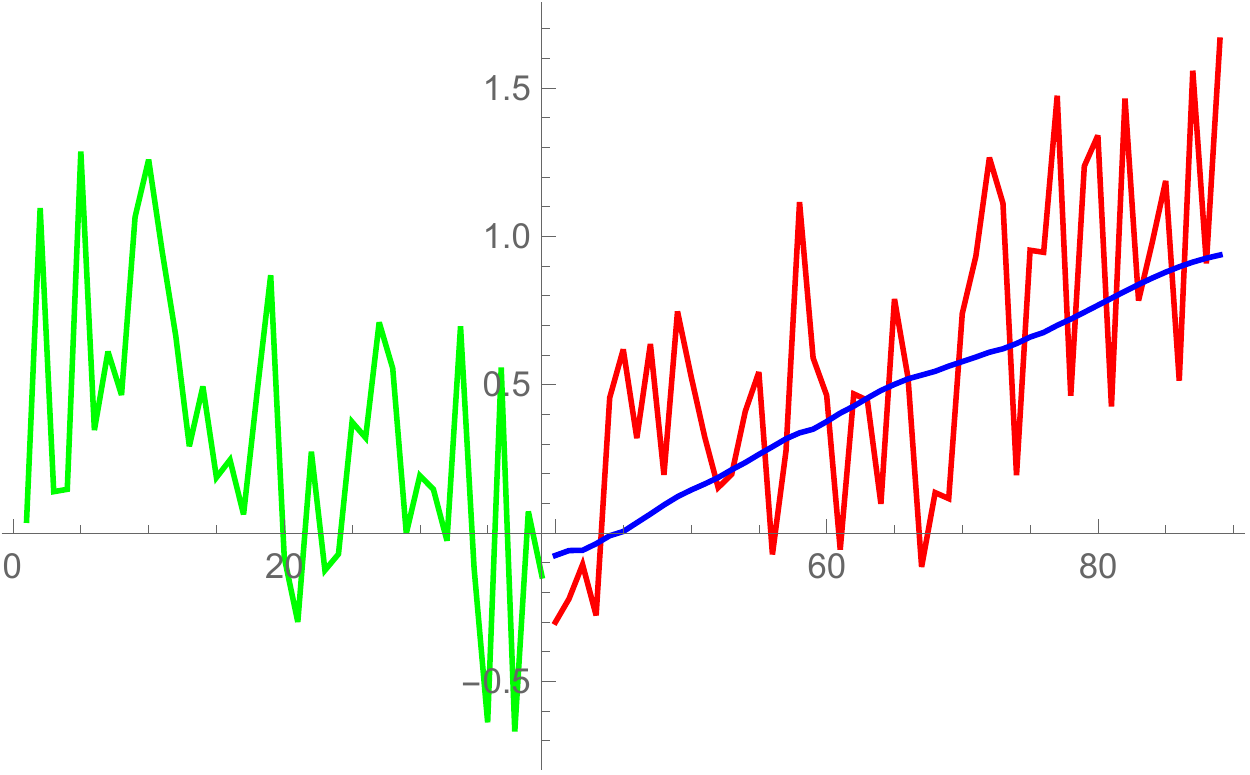,width=7cm}
\\
 ({\bf c}) & ({\bf d})
\end{tabular}
\caption{The input segment of a sequence (green) 
with ({\bf a}) $a_i=0$,  ({\bf b}) $a_i=0.15$,  ({\bf c}) $a_i=0.40$,  ({\bf d}) $a_i=0.75$ 
of triangle wave, the subsequent segment of 
the data sequence (red)
and predicted dynamics (blue) using LSTM layer network with $20$  neurons trained 
using data sets with noise amplitude $a_0=0.15$.
}
\label{Fig2a}
\end{center}
\end{figure}

It should be noted that networks with LSTM recurrent layer demonstrate
better training stability and higher prediction quality compared to the ones with basic layer
as shown in Fig. \ref{Fig2a}.

\subsection{Prediction results for two dimensional trajectories}

\begin{figure}[h!]
\begin{center}
\begin{tabular}{cc}
\psfig{figure=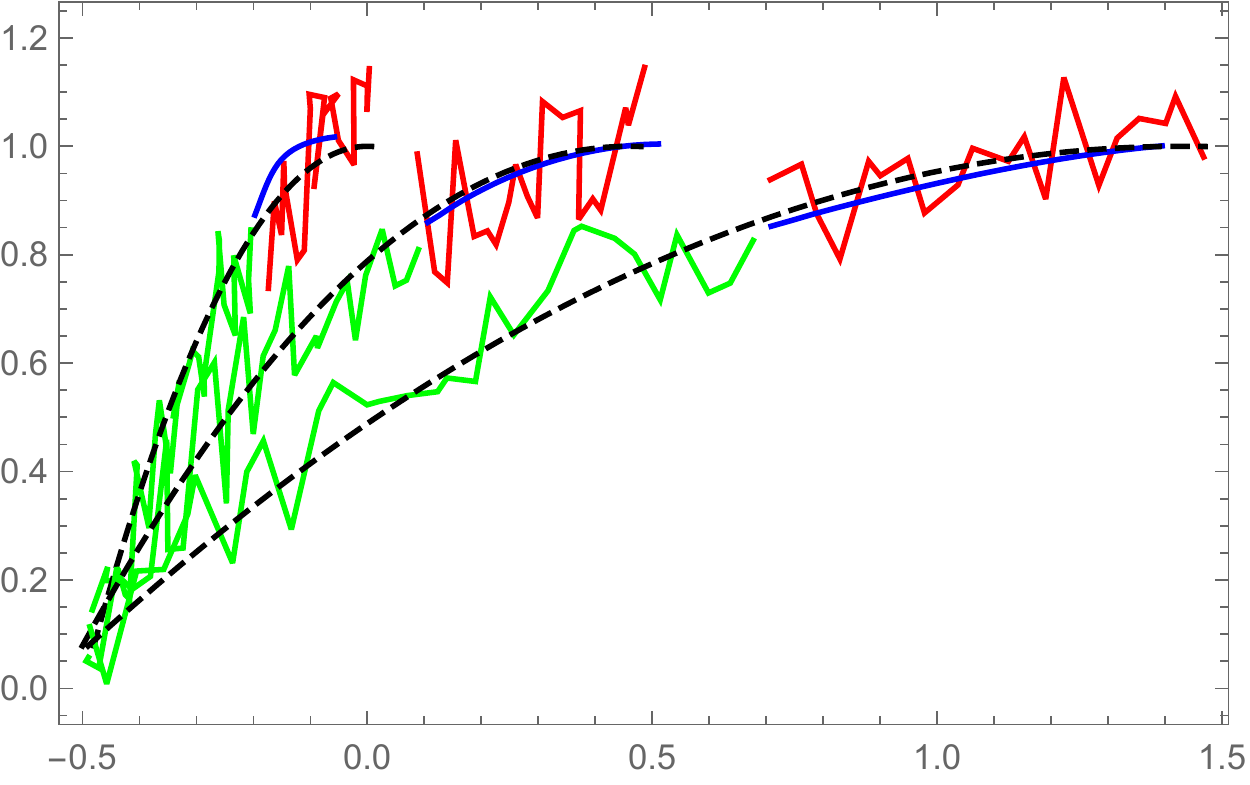,width=7.5cm} 
&
\psfig{figure=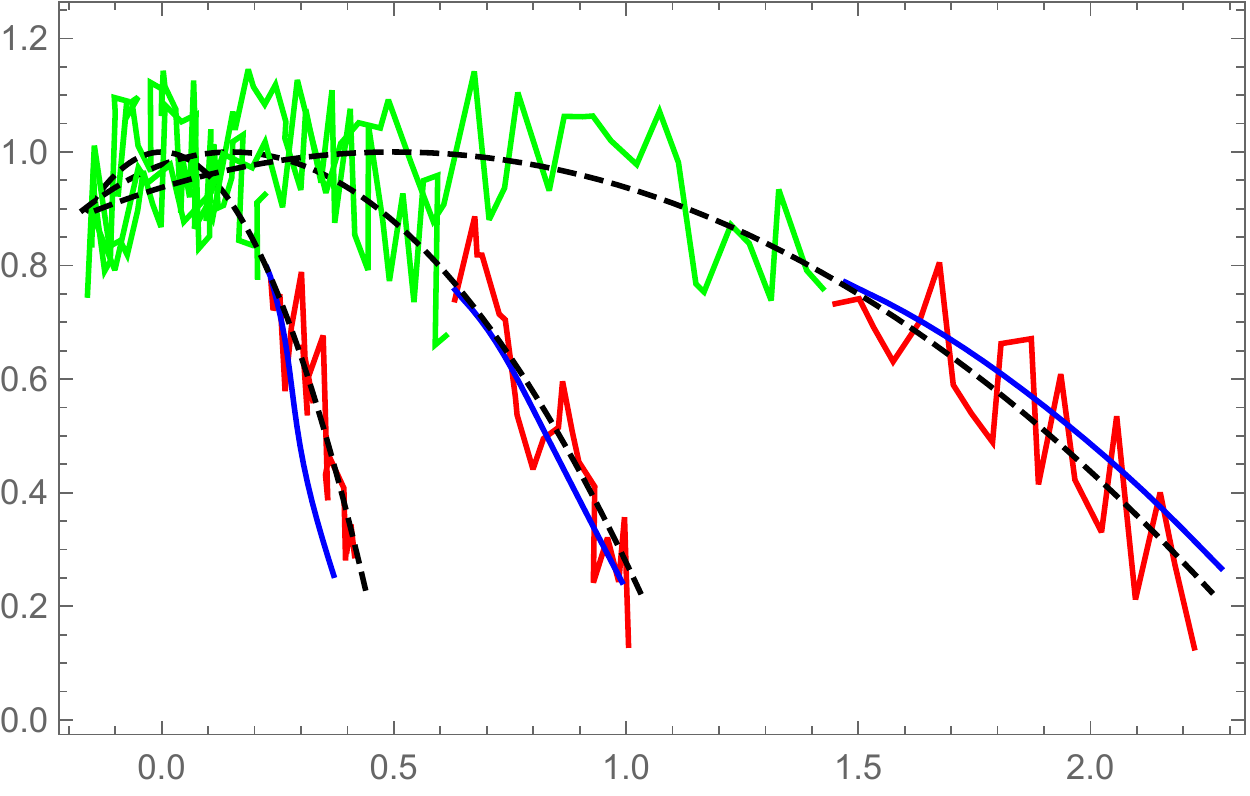,width=7.5cm}
\\
({\bf a}) & ({\bf b}) \\
\psfig{figure=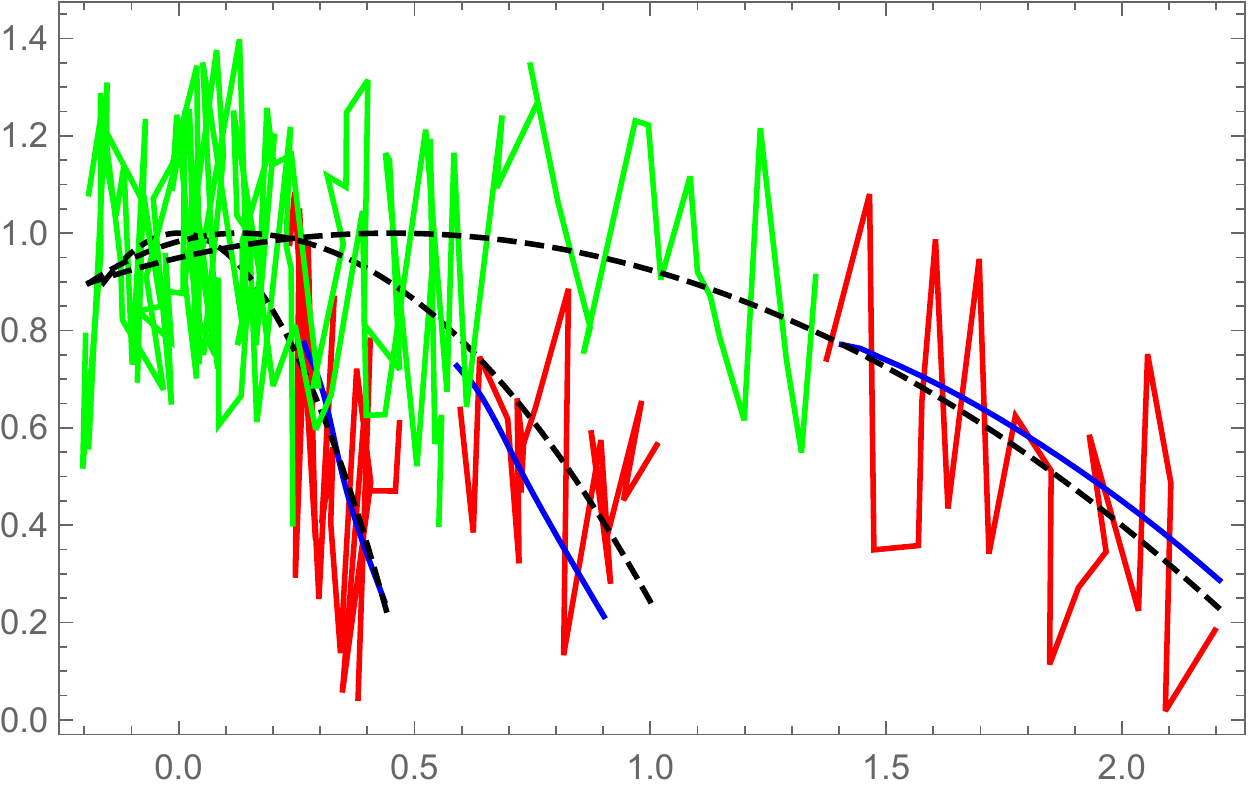,width=7.5cm}
&
\psfig{figure=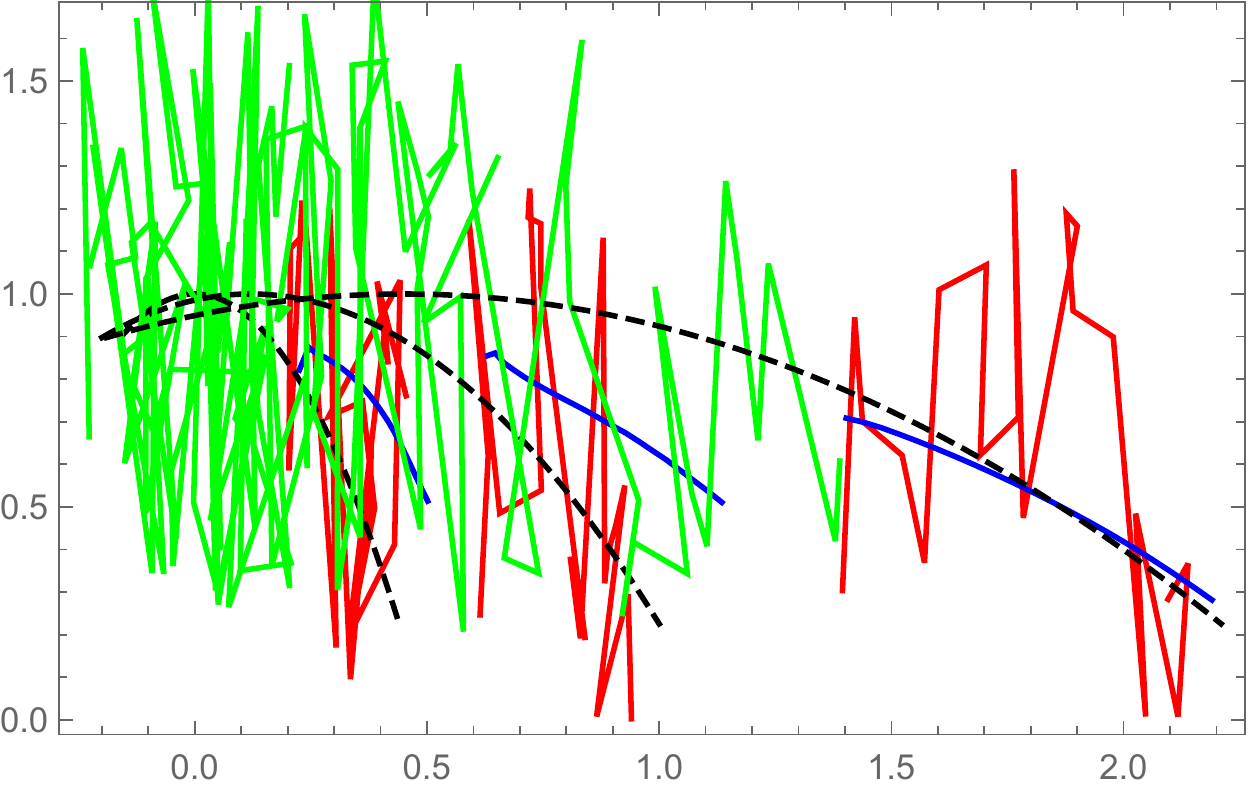,width=7.5cm}
\\
({\bf c}) & ({\bf d})
\end{tabular}
\caption{The input segment of a sequence (green) 
with ({\bf a,b}) $a_i=0.15$, ({\bf c}) $a_i=0.40$,  ({\bf d}) $a_i=0.75$ 
of three parabolas, the subsequent segment of 
the data sequence (red)
and predicted dynamics (blue) by LSTM layer network with $20$  neurons trained 
on data sets with noise amplitude $a_0=0.15$. For comparison smooth unperturbed dynamics 
is shown by dashed black curve.
}
\label{Fig4}
\end{center}
\end{figure}

\begin{figure}[h!]
\begin{center}
\begin{tabular}{cc}
\psfig{figure=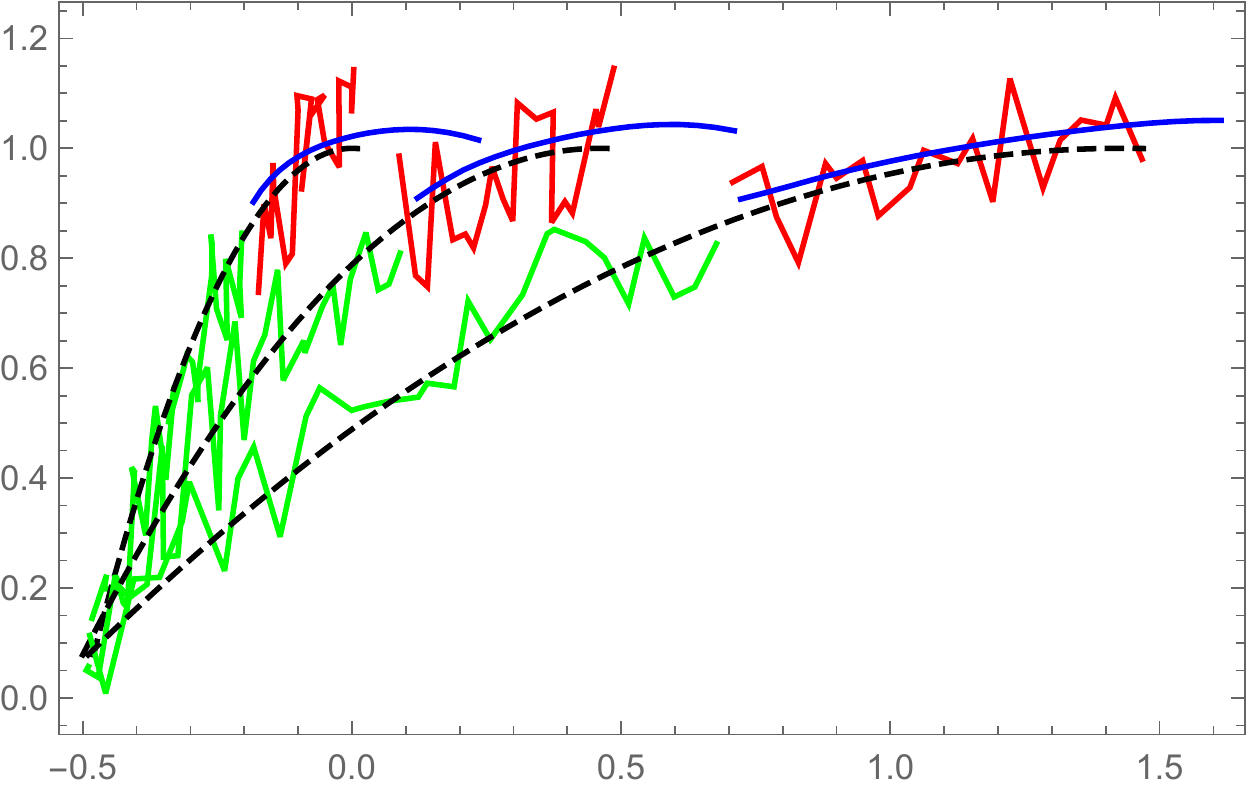,width=7.5cm} 
&
\psfig{figure=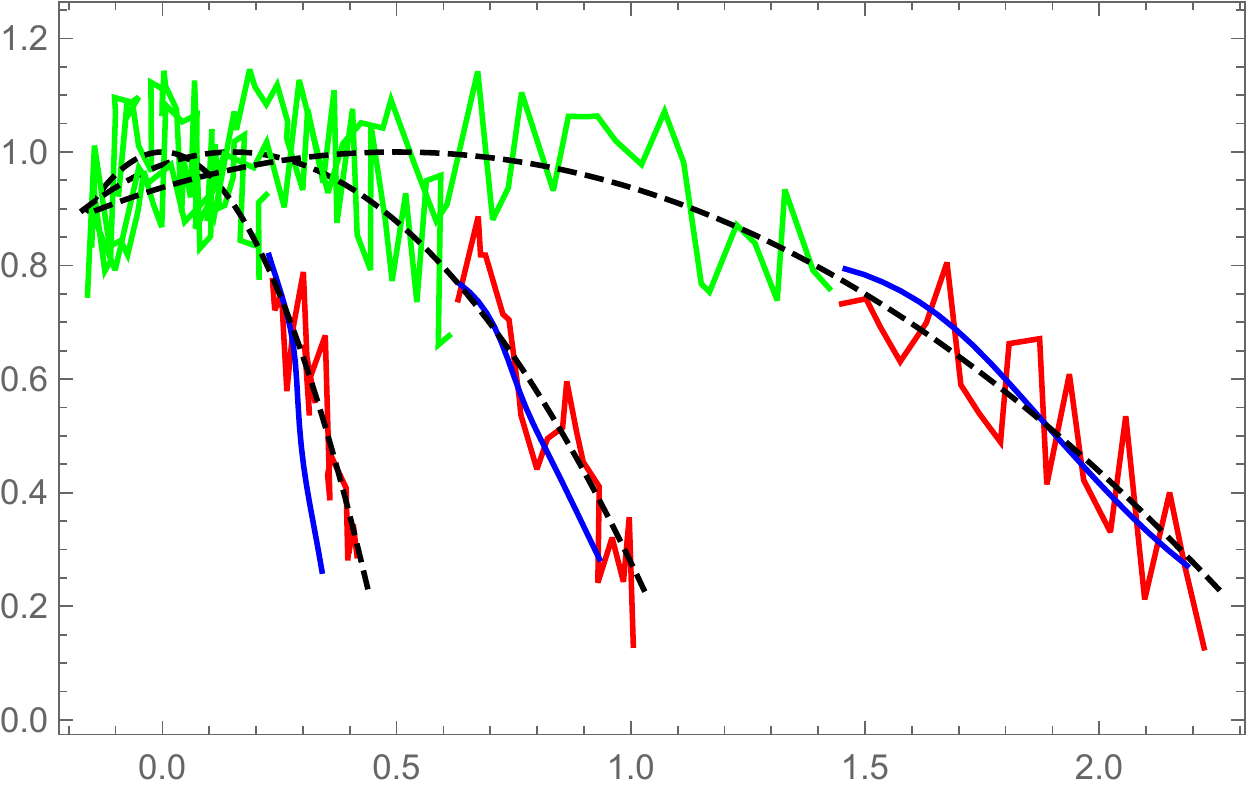,width=7.5cm}
\\
({\bf a}) & ({\bf b}) \\
\psfig{figure=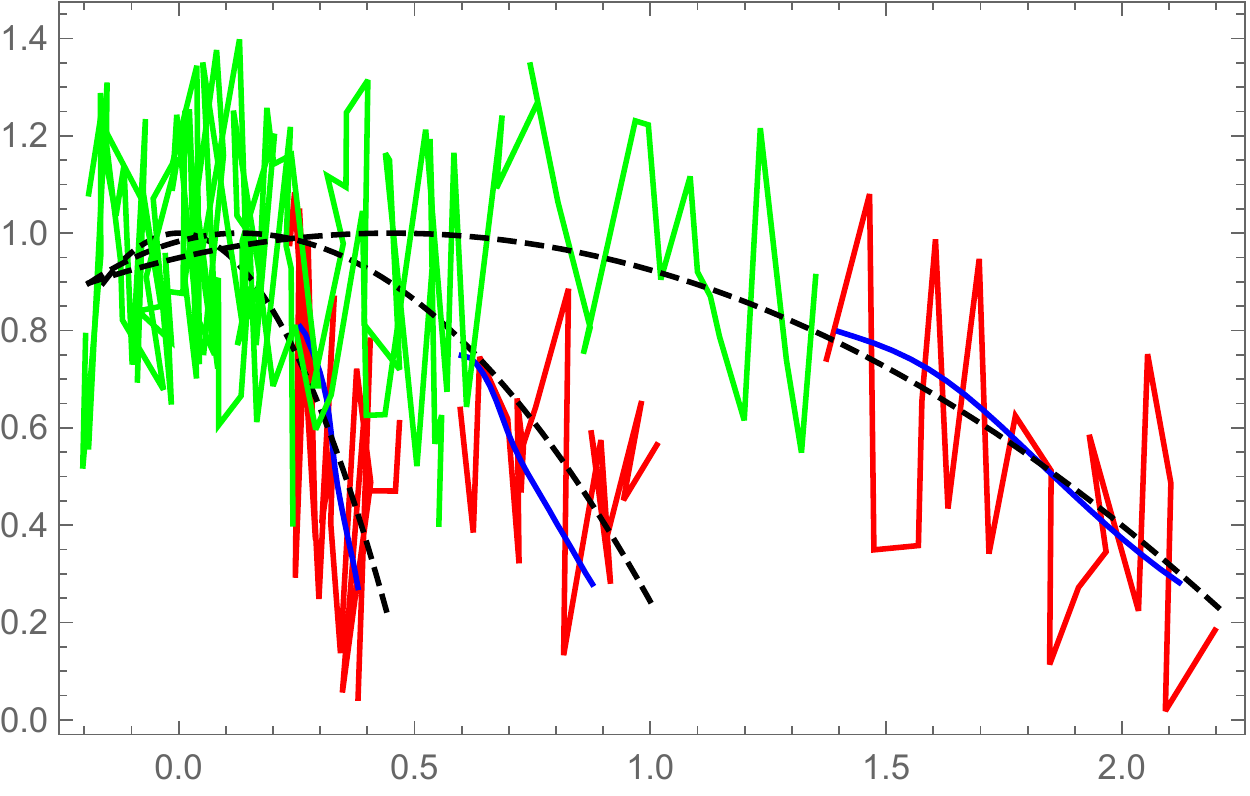,width=7.5cm}
&
\psfig{figure=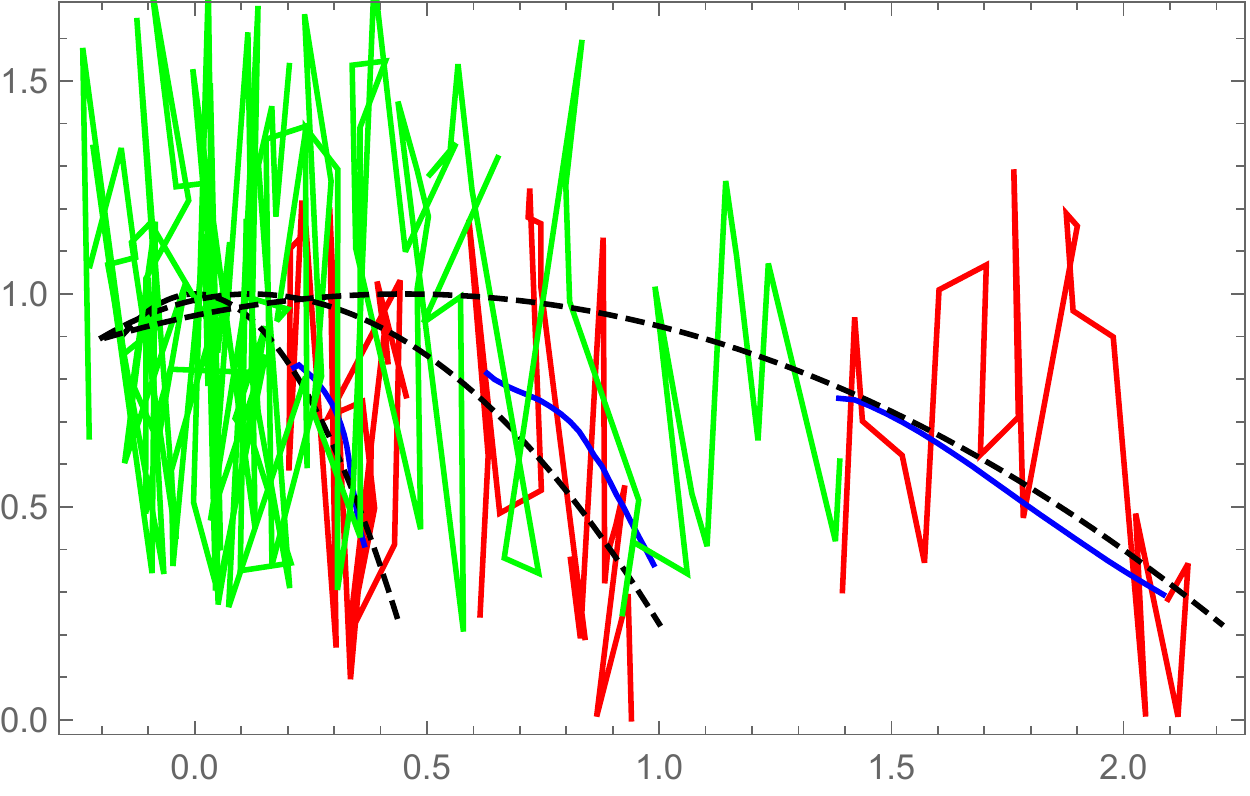,width=7.5cm}
\\
({\bf c}) & ({\bf d})
\end{tabular}
\caption{The input segment of a sequence (green) 
with ({\bf a,b}) $a_i=0.15$, ({\bf c}) $a_i=0.40$,  ({\bf d}) $a_i=0.75$ 
of three parabolas, the subsequent segment of 
the data sequence (red)
and predicted dynamics (blue) by LSTM layer network with $20$  neurons trained 
on data sets with noise amplitude $a_0=0.4$. For comparison smooth unperturbed dynamics 
is shown by dashed black curve.
}
\label{Fig4a}
\end{center}
\end{figure}

As in 1D case the PRN trained on the data sets 
based on the smooth functions fails to predict 
for noisy inputs. When the training data contains noise
the networks appear to successfully generate quite smooth
segments of all three parabolas 
(both ascending and descending range of curve). 
As shown in Fig. \ref{Fig4}a ($a_0=0.15$) the prediction quality
on the ascending segment improves for increasing trajectory range. 
On the descending segment dynamics is predicted quite well 
for the input noise comparable to the training noise level 
(Fig. \ref{Fig4}b,c) but for significantly larger noise the network
fails to predict short and intermediate size parabolas (Fig. \ref{Fig4}d).
To remedy this weakness, we train the network for higher value $a_0=0.4$ of the 
noise component $\bm \xi$ and observe slight improvement of the 
prediction quality for large noise input (Fig. \ref{Fig4a}a).

\section{Why do noisy inputs lead to smooth predictions?}

In the preceding section we consider the influence of
noise in the training data as well as in the network input 
sequences on the network prediction and show that 
a trained network converts noisy input to
smooth predictions.
As mentioned above the numerical simulations 
performed in \cite{Rub2020a,Rub2020b,Rub2021a} demonstrate
that the observed PRN behavior is independent of
its size, architecture type 
as well as the predictive algorithm choice -- 
either the "moving window", the "expanding window" or memoryless
procedure introduced in \cite{Rub2020a}.
It is reasonable to assume that the smoothness
of the predicted dynamics is somehow related
to the training procedure. 

Before turning to the training procedure analysis
we consider how the noise component in the elements 
$\bm x_i$ of the 
input sequence $\bm X$ influences the dynamics
of the network inner states $\bm s_i$ \cite{Rub2020a}.

\subsection{RNN does not filter out input sequence noise component}
When an input sequence $\bm X = \{\bm x_i\}$ is supplied to a network
a sequence of the inner states $\bm S= \{\bm s_i\}$ is generated 
by recursive application of the rule (\ref{map}). 
Consider a case of the noisy input with $\bm x_i = \bm f_i + a \bm \xi_i$
where $a$ is the noise amplitude.
Compute $\bm s_1$ 
$$
\bm s_1 = \bm F(\bm f_1+a\bm \xi_1,\bm s_0) 
\approx 
 \bm F(\bm f_1,\bm s_0) +
a \bm J_{x1} \cdot \bm \xi_1 =
\hat{\bm s}_1 + a \bm \sigma_1,
$$
where 
$\bm J_{x1} = \partial  \bm F(\bm f_1,\bm s_0) /\partial \bm x,$ and
$\hat{\bm s}_1 =  \bm F(\bm f_1,\bm s_0) $
is the inner state value computed for the noiseless input $\bm f_1$.
We observe that $\bm s_1$ has a noisy component proportional to the 
input noise amplitude $a$. Turn to $\bm s_2$ evaluation
$$
\bm s_2 = \bm F(\bm f_2+a\bm \xi_2,\bm s_1) = 
\bm F(\bm f_2+a\bm \xi_2,\hat{\bm s}_1 + a \bm \sigma_1) 
\approx
\bm F(\bm f_2,\hat{\bm s}_1)+
a \bm J_{x2} \cdot \bm \xi_2+
a \bm J_{s1} \cdot \bm \sigma_1 = 
\hat{\bm s}_2 + a \bm \sigma_2,
$$
where 
$\bm J_{x2} = \partial  \bm F(\bm f_2,\hat{\bm s}_1) /\partial \bm x,$ 
$\bm J_{s1} =\partial  \bm F(\bm f_2,\hat{\bm s}_1) /\partial \bm s,$
and we obtain for the noiseless and noisy contributions to $\bm s_2$
the expressions
$$
\hat{\bm s}_2 = \bm F(\bm f_2,\hat{\bm s}_1),
\quad
\bm \sigma_2 =  \bm J_{x2} \cdot \bm \xi_2+
\bm J_{s1} \cdot \bm \sigma_1.
$$
Repeating the procedure we obtain for $\bm s_i=\hat{\bm s}_i + a \bm \sigma_i$, with 
$\hat{\bm s}_i = \bm F(\bm f_i,\hat{\bm s}_{i-1}),$ and
\bea
&&
\bm \sigma_i =  \bm J_{xi} \cdot \bm \xi_i+
\bm J_{s,i-1} \cdot \bm \sigma_{i-1},
\label{sigma_dyn}
\\
&&
\bm J_{xi} = \partial  \bm F(\bm f_i,\hat{\bm s}_{i-1})/\partial \bm x,
\quad\quad
\bm J_{s,i-1} = \partial  \bm F(\bm f_i,\hat{\bm s}_{i-1})/\partial \bm s,
\label{sigma_matr}
\eea
and thus demonstrate that the inner states of the network 
have noise component proportional to that of in the input sequence.

The dynamics of the noise component $\bm \sigma$ in the linear 
approximation defined in (\ref{sigma_dyn}) strongly depends on the 
spectrum of the square matrix $\bm  J_{s,i-1}$ 
in (\ref{sigma_matr}). For any square matrix $\bm W_s$ one can find its eigenvalues $\lambda_j$; if 
$|\lambda_j| < 1$ for all $j$ the linear transformation defined by the  matrix $\bm W_s$
is contractive, i.e., $|\bm W_s \cdot \bm u| < |\bm u|$ for any vector $\bm u$.

Note that 
$\bm  J_{s,i-1}$ in (\ref{sigma_matr}) is obtained 
as partial derivative of the nonlinear vector function $\bm F$ with parameters 
defined in the result of 
PRN training and there is no guarantee that it satisfies the contraction 
conditions. Even if it is the case and we have $|\bm J_{s,i-1} \cdot \bm \sigma_{i-1}| < |\bm \sigma_{i-1}|$,
the first term in r.h.s. of (\ref{sigma_dyn}) cannot be neglected as 
$\bm \xi_{i}$ represents a {\it random} noise contribution to the point $\bm x_i$.
This means that the network does not filter out noise present in the input sequence $\bm X$.
In the multi-layer PRN the output of a given (inner) layer plays a role 
of the input to the next layer, so that the noise component 
propagates through all network layers without decay.

As the predicted value $\bar{\bm x}$ depends linearly
on the last inner state $\bm s_m$ one expects that
the predicted value $\bar{\bm x}$ would also have a noisy component. 
It is true for an untrained network but training makes this expectation invalid.
To understand this fact better we have to consider different 
training procedures in more details.

\subsection{"Moving window" prediction procedure}

\begin{figure}[h!]
\begin{center}
\begin{tabular}{cc}
\psfig{figure=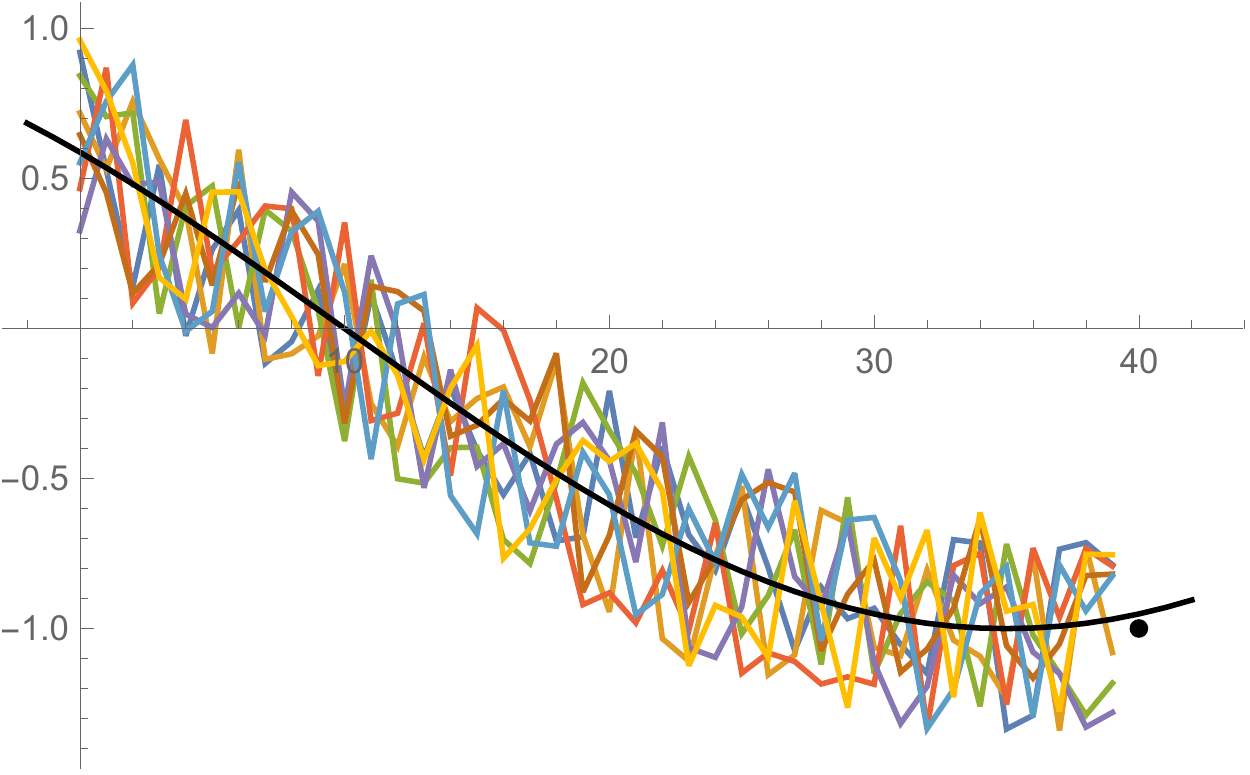,width=7.0cm} 
&
\psfig{figure=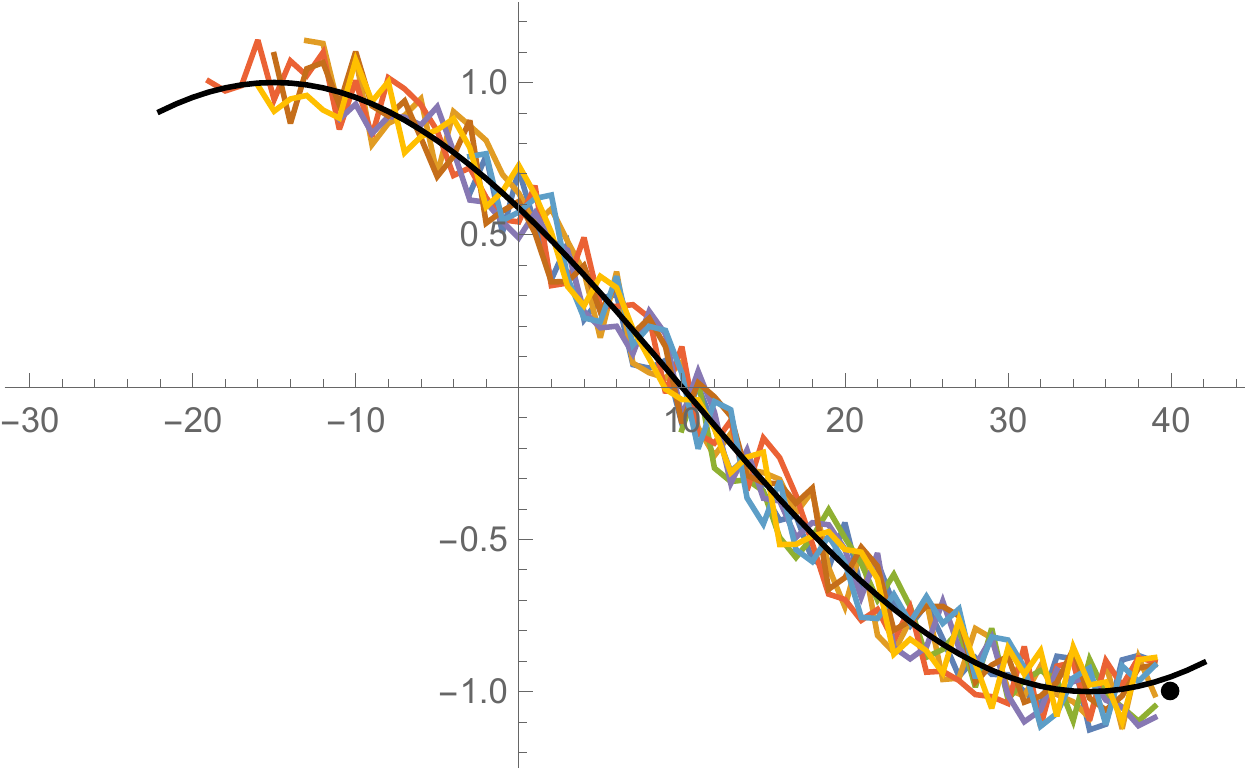,width=7.0cm}
\\
({\bf a}) & ({\bf b}) 
\end{tabular}
\caption{Averaging of the predicted point during
"seq-to-one" training procedure with
({\bf a}) fixed and ({\bf b}) variable input sequence length. 
The noise amplitude of eight segments of the training data sets is $a_0=0.15$. 
Black dots denote an average of the predicted value and
black solid curve represents a segment of the actual sine wave trajectory.
}
\label{Fig5}
\end{center}
\end{figure}

Consider the "seq-to-one" training procedure
with the sequences of a fixed length 
that corresponds to the MW prediction.
Each time when the sequence of length $m$
has the same initial point $j_s$ the input values
read $\bm g_j = \bm f_j + a_i \bm \xi_j,j_s \le  j \le j_s+m$,
where $\bm f_j$ are the same but $\bm \xi_j$ are the 
random vectors with zero average. The same observation 
is valid for the point with $j=j_s+m+1$ to be predicted --
$\bm g_{j_s+m+1} = \bm f_{j_s+m+1} + a_i \bm \xi_{j_s+m+1}$.
As during the training procedure this segment is used repeatedly with changing 
noise component
the network eventually is trained to attempt prediction of an average of these points
$\langle \bm g_{j_s+m+1} \rangle \approx \bm f_{j_s+m+1}$ as 
the noise component averages out (Fig. \ref{Fig5}a).
When the number of points $n_t$ representing a trajectory is much smaller than 
the number of data sets $N_d$ used in training procedure covering the whole trajectory,
each point of the trajectory would be visited approximately $N_d/n_t \gg 1$ times
and the averaging should be quite effective.
As the result the training procedure for every input sequence effectively
forces the network to predict {\it nearly all} points in
a very small vicinity of the actual curve. 
It implies that the deviation of these points from the actual trajectory
is much smaller than in the input noisy sequence and the predicted trajectory
looks smooth.
In other words when the original noisy sequence $\bm X$ is 
supplied to the trained network the latter tries to generate the 
last inner state $\bm s_m$ that is very close to its noiseless
counterpart $\bm s_m \approx \hat{\bm s}_m$ (Fig. \ref{Fig5}).

It is instructive to underline important difference between periodic
(spatially infinite) and finite size trajectories represented by $N$ points. On the latter we have 
one special point qualitatively different from all other points -- 
this is the trajectory last $N$-th point. To predict this point, one has to use for training the 
segments ending at $(N-1)$-th point. An average number of points
in such segments is smaller than $n_t$ but the number of these segments
is larger than $N_d$. This means that the last point is predicted with higher accuracy
compared to all other points of the trajectory (Fig. \ref{Fig6}b).

Thus the "seq-to-one" training paradigm forces 
a network to produce a trajectory that is close to underlying 
noise-free dynamics.



Consider rounds of the prediction using the memoryless algorithm that
produces a predicted trajectory coinciding with that of
made by the original MW algorithm.

\underline{Round 1.}
The input $\bm X^1 = \{\bm x_j\},\ 1 \le i \le m$ 
consists of the elements of the original input sequence
and has no predicted elements. As the predicted value
$\bar{\bm x}^1 = \bar{\bm x}_{m+1} \approx \bm f_{m+1}$
is in the close vicinity of the actual trajectory we deduce that
$\bm s_m^1 \approx \hat{\bm s}_m^1$ such that 
$\bm f_{m+1} = \bm L(\hat{\bm s}_m^1) = \bm W \cdot \hat{\bm s}_m^1 + \bm B$.

\underline{Round 2.}
For the second prediction round we form a new input 
$\bm X^2 = \{\bm x_2,\ldots,\bm x_m, \bar{\bm x}_{m+1}\}$
and find 
$$
\bm s_m^2 = \bm F(\bar{\bm x}_{m+1}, \bm s_{m}^{1})
\approx \bm F(\bm f_{m+1},\hat{\bm s}_m^1) =
\hat{\bm s}_m^2.
$$
Use it to predict 
$\bar{\bm x}_{m+2}=\bm L(\bm s_m^2)\approx \bm L(\hat{\bm s}_m^2)$
that leads to a noiseless predicted value
$\bar{\bm x}_{m+2}\approx\bm f_{m+2}$.

\underline{Round 3.}
Repeat the above procedure to obtain
$\bm s_m^3 = \bm F(\bar{\bm x}_{m+2}, \bm s_{m}^{2})
\approx \bm F(\bm f_{m+2},\hat{\bm s}_m^2) =
\hat{\bm s}_m^3$,
and we again find $\bar{\bm x}_{m+3}\approx\bm f_{m+3}$.

One can see that at $k$-th prediction round we have
\be
\bm s_m^k = \bm F(\bar{\bm x}_{m+k-1}, \bm s_{m}^{k-1})
\approx \bm F(\bm f_{m+k-1},\hat{\bm s}_m^{k-1}) =
\hat{\bm s}_m^k,
\quad
\bar{\bm x}_{m+k}\approx\bm f_{m+k}.
\label{MWsmooth}
\ee

\subsection{"Expanding window" prediction procedure}

\begin{figure}[h!]
\begin{center}
\begin{tabular}{cc}
\psfig{figure=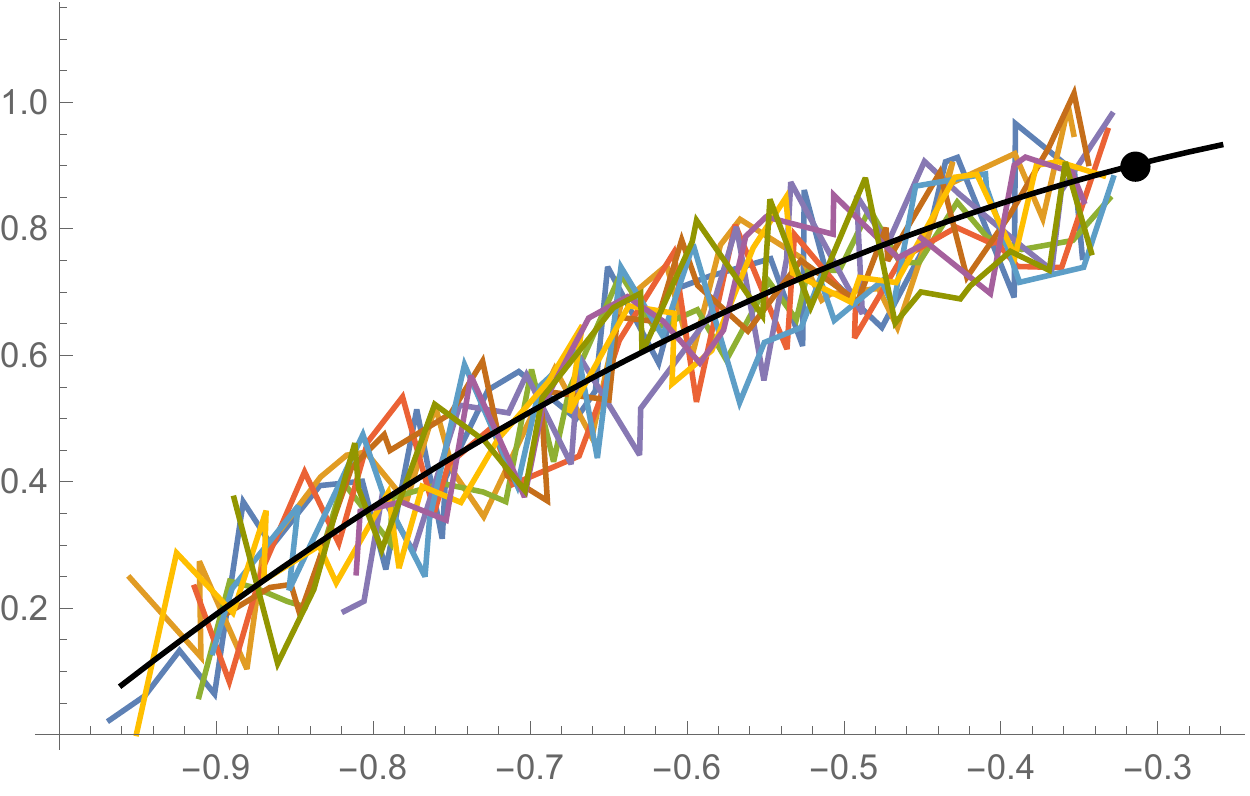,width=7.0cm} 
&
\psfig{figure=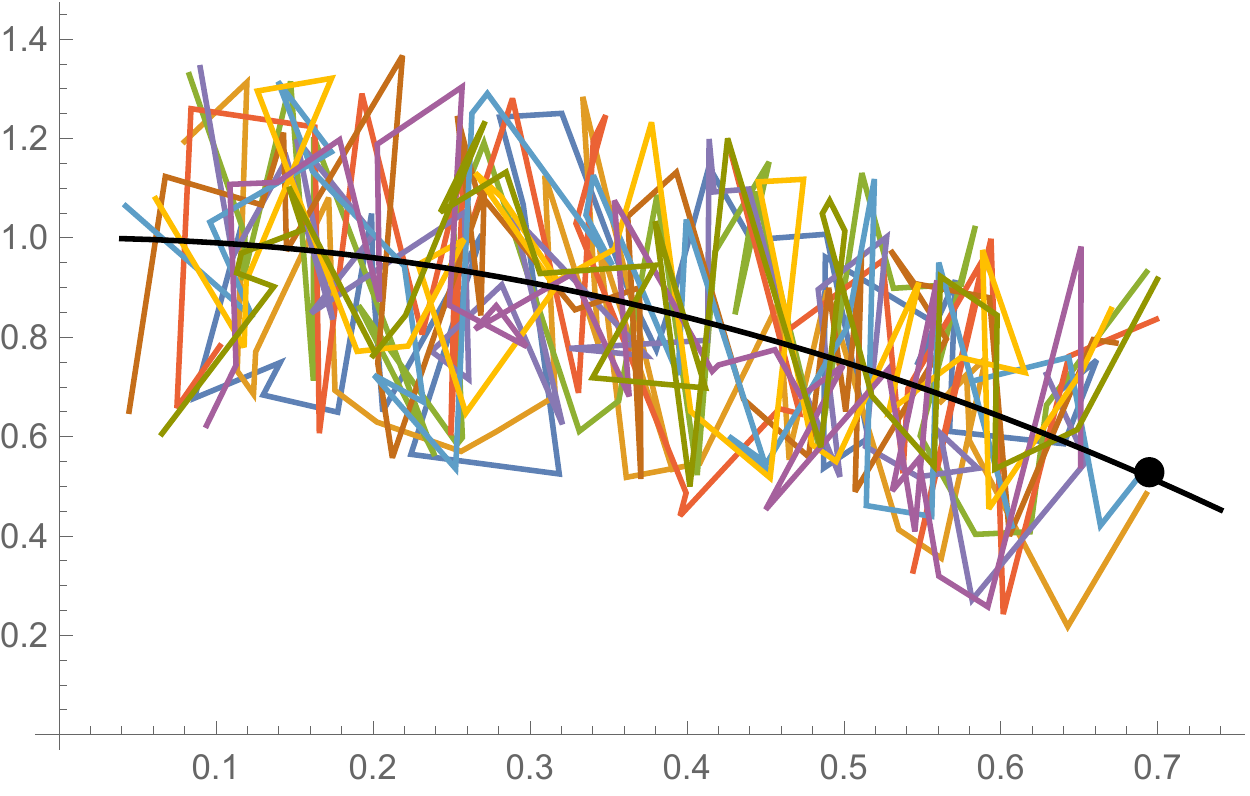,width=7.0cm}
\\
({\bf a}) & ({\bf b}) 
\end{tabular}
\caption{Averaging of the predicted point during
"seq-to-one" training procedure with  variable input sequence length in the 
({\bf a}) ascending and ({\bf b}) descending segment of parabolic trajectory. 
The number and noise amplitude of the training segments is 
(a) $10$ segments with $a_0=0.15$ and (b) $20$ segments with $a_0=0.4$. 
Black dots denote an average of the predicted value and
black solid curve represents the actual trajectory segment.
}
\label{Fig6}
\end{center}
\end{figure}

Now turn to the training procedure using segments of variable 
length corresponds to the EW prediction. Consider all training segments
having the last point $\bm g_{j_s+m} = \bm f_{j_s+m} + a_i \bm \xi_{j_s+m}$.
The point for the prediction for all such segments has 
$j=j_s+m+1$. As we note above the noise component of all these 
points should average out and it again predicts 
$\langle \bm g_{j_s+m+1} \rangle \approx \bm f_{j_s+m+1}$. 
(Fig. \ref{Fig5}b, \ref{Fig6}).

\underline{Round 1.}
The input $\bm X^1 = \{\bm x_j\},\ 1 \le i \le m$ 
consists of the elements of the original input sequence
and has no predicted elements. As the predicted value
$\bar{\bm x}^1 = \bar{\bm x}_{m+1} \approx \bm f_{m+1}$
is in the close vicinity of the actual trajectory we deduce that
$\bm s_m^1 \approx \hat{\bm s}_m^1$ such that 
$\bm f_{m+1} = \bm L(\hat{\bm s}_m^1) = \bm W \cdot \hat{\bm s}_m^1 + \bm B$.

\underline{Round 2.}
For the second prediction round we have to form a new input  
$\bm X^2$ with $m+1$ elements
by adding the predicted value
$\bar{\bm x}_{m+1}$ to $\bm X^1$ and producing
$\bm X^2 = \{\bm x_1,\bm x_2,\ldots,\bm x_m, \bar{\bm x}_{m+1}\}$.
Note that the last element of $\bm X^2$ has much lower noise amplitude compared to
all preceding terms.
Feeding this sequence into the network we produce a sequence
of the inner states $\bm s_i^2$ using (\ref{map}) as
$\bm s_i^2 = \bm F(\bm x_i^2,\bm s_{i-1}^2)$.
Evaluate the last term $\bm s_{m+1}^2=\bm F(\bm x_{m+1}^2,\bm s_{m}^2)$
and note that $\bm x_{m+1}^2 = \bar{\bm x}_{m+1} \approx \bm f_{m+1}$. 
As $\bm s_{m}^2 \approx \hat{\bm s}_m^1$ 
we find both arguments of $\bm F$ correspond to noiseless values leading to
$$
\bm s_{m+1}^2 \approx \bm F(\bm f_{m+1},\hat{\bm s}_m^1)  = 
\hat{\bm s}_{m+1},
\quad
\bar{\bm x}_{m+2} = \bm L(\bm s_{m+1}^2) \approx \bm f_{m+2}.
$$
and the predicted value $\bar{\bm x}_{m+2}$ is also close to the 
actual trajectory.

\underline{Round 3.}
In the next round we create an input $\bm X^3$ with two last values $\bm x_{m+2}^3 \approx \bm f_{m+2}$ 
and $\bm x_{m+1}^3\approx \bm f_{m+1}$ close to the 
actual trajectory. We have 
$\bm s_{m+1}^3 \approx \hat{\bm s}_{m+1}$ so that 
$\bm s_{m+2}^3 =  \bm F(\bm x_{m+2}^3,\bm s_{m+1}^3) \approx 
\bm F(\bm f_{m+2},\hat{\bm s}_{m+1})= \hat{\bm s}_{m+2}$ and we 
obtain 
$\bar{\bm x}_{m+3} = \bm L(\bm s_{m+2}^3) \approx \bm f_{m+3}$.

Repeating the prediction rounds in similar manner we arrive at the 
general $k$-th term of the predicted sequence 
\be
\bm s_{m+k-1}^k =  \bm F(\bm x_{m+k-1}^k,\bm s_{m+k-2}^k) \approx 
\bm F(\bm f_{m+k-1},\hat{\bm s}_{m+k-2})= \hat{\bm s}_{m+k-1},
\
\bar{\bm x}_{m+k} = \bm L(\bm s_{m+k-1}^k) \approx \bm f_{m+k}.
\label{EWsmooth}
\ee

The results (\ref{MWsmooth},\ref{EWsmooth}) demonstrate 
that both a sequence of the predicted values $\bar{\bm x}_{m+k}$ and 
a sequence of the corresponding network inner states $\bm s_m^k$ represent
smooth trajectories in their respective phase space.



\section{Recurrent network architecture generalizations}
In the preceding sections we consider simple PRN with inner
dynamics described by (\ref{map}) trained to predict a single point of 
trajectory (``seq-to-one'' networks) that follows the input sequence. We demonstrate and explain
an unexpected ability of such networks being well trained on noisy sequences
representing an apparent trajectory to predict actual smooth trajectory
It is instructive to verify whether this property persists in networks
with more general architectures, namely ``seq-to-seq'' networks.

\begin{figure}[h!]
\begin{center}
\begin{tabular}{cc}
\psfig{figure=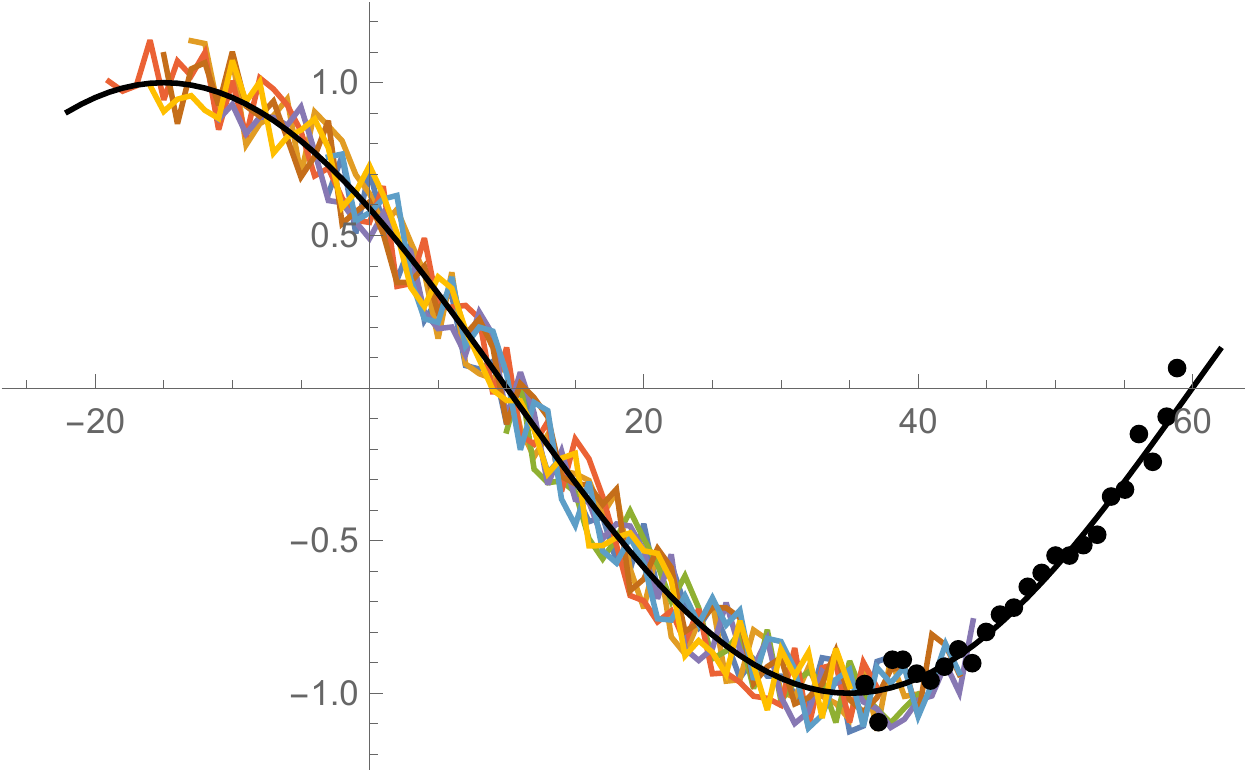,width=7.0cm} 
&
\psfig{figure=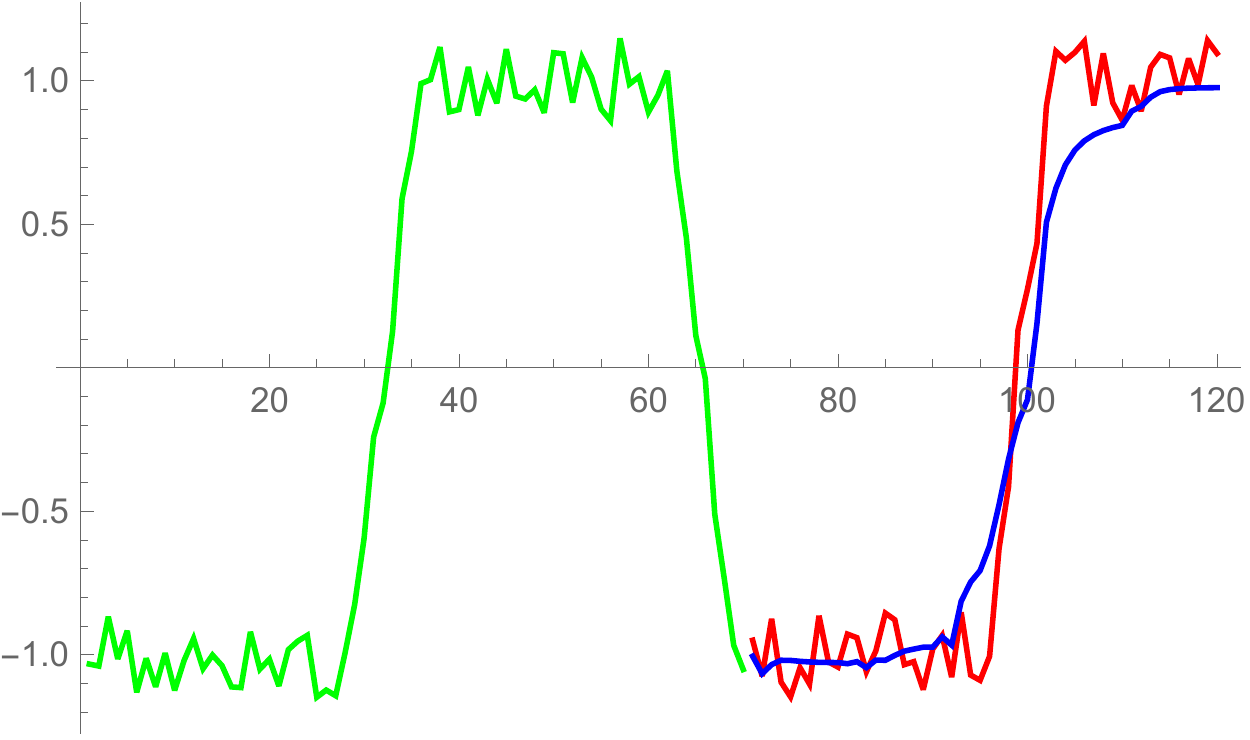,width=7.0cm}
\\
({\bf a}) & ({\bf b}) 
\end{tabular}
\caption{
({\bf a}) Averaging of the predicted points for
"seq-to-seq" training with variable input.
The noise amplitude of eight segments of the training data sets is $a_0=0.15$. 
Black dots denote an average of $l=15$ values in the predicted sequence and
black solid curve represents a segment of the actual sine wave trajectory.
({\bf b}) Comparison of the ground truth continuation (red) of the input noisy phase modulated 
trapezoid wave sequence (green)
to the predictions computed by MW algorithm (solid blue)
in the network with the total number of neurons $n=50$.
The length of the input sequence $m=70$ and the predicted sequence size is $kl=40$ with
$l=10$.
}
\label{Fig7}
\end{center}
\end{figure}

``Seq-to-seq'' architecture uses the 
input sequence of fixed or variable length while the 
predicted sequence has a fixed length $l>1$. 
It uses a training procedure \cite{seq2seq2014} with
the last element $\bm s_m$ of the inner state sequence 
copied $p$ times and this new sequence is 
submitted to another PRN that eventually generates
the desired output sequence 
$\{\bar{\bm x}_{m+j}\},\ 1 \le j \le l$.
During the training all predicted sequences having the 
same initial point or those that overlap significantly
contribute to averaging that effectively allows to 
predict the fixed size segment of the trajectory
close to the actual one. As the
complete overlapping of the predicted segments 
is relatively rare the prediction curve smoothness
is expected to be lower than in case of the  
``seq-to-one'' training algorithm (Fig. \ref{Fig7}a). 
It is confirmed by
numerical simulations in \cite{Rub2021a} of prediction
of phase modulated trapezoid wave (Fig. \ref{Fig7}b)
that also shows that the ``roughness'' of the predicted curve
is still smaller than that of the input sequence.

\section{Noiseless prediction of time series in biology context}

In our opinion the "built-in" ability of RNNs to generate smooth
prediction dynamics plays very important 
role in biological evolution of mobile organisms. 
Any mobile organism that receives and processes information about 
position and dynamics of other moving objects and immobile obstacles
in its vicinity should be able at least to avoid collisions
with these objects to increase its survival probability. 
To this end the organism should have an
ability to predict trajectories in space.
In most cases actual trajectories have 
negligible low noise amplitude but the corresponding apparent ones
processed by the organism predictive network might have
much higher noise levels. The origin of this noise
includes motion of the organism itself and 
errors in the sensors detecting trajectories.

It is important to underline that ubiquity of noisy 
sequences encoding smooth trajectories
forces PRNs to learn to predict based on
noisy input data sets.
The same time it is reasonable to assume that the organism {\it has to predict}
the actual smooth trajectory -- if otherwise a noisy trajectory
is predicted the error of prediction would be too large in order
to generate a required reaction of the organism.
It looks impossible to 
resolve the apparent contradiction between a perturbed noisy input
trajectory and a desired smooth predicted dynamics that emerges
in biological context.
But it appears that this requirement is satisfied nearly 
automatically as a default feature of PRNs of sufficient size
trained on the noisy data sets that arise {\it naturally}
in biology.

Turning to prediction of specific trajectory types
we start with computationally preferable infinite one-dimensional
periodic sine and triangle waves. The simulations confirm that 
indeed PRNs trained on noisy datasets successfully predict 
basic smooth curves. Unfortunately, perfectly periodic 
in space and time trajectories 
are nearly absent in nature. Instead finite trajectories
are omnipresent and thus their prediction is important for survival.
One class of such trajectories includes parabola describing
with high accuracy
passive motion of a (small) solid object under gravity force.
The prediction of an ascending segment of parabola is a valuable
trait for getting fruit hanging from a free. The same time accurate guess of the 
final segment of a descending tail of parabola is important
to evade being hit by a flying object.

The parabola has two important parameters -- the maximal height $h$ 
and the range (distance between the initial and end points) $b$, 
their ratio $h/b$ is critical for prediction process. Essentially
we argue that both extreme cases $h/b \ll 1$ and $h/b \gg 1$ represent
curves that nearly impossible to predict. We choose three parabolas
with $h/b = 1, 1/2, 1/4$ and fixed $h=1$ as typical examples
of predictable trajectories and trained PRNs designed to 
"recognize" and reproduce all three curves.
We show that for noisy input sequence representing these essentially two-dimensional curves
PRNs can predict smooth continuation
that follows the unperturbed path of a moving object.


The main result of this manuscript 
also may be applicable in more general context of
biological evolution. This evolution is viewed as a process of learning
by a network that tries to predict 
dynamics of the environment
\cite{VanchurinPNAS2022}.
In this approach an organism fitness increases 
when it can successfully predict the environment 
variables dynamics, {\it i.e.},
when the predictive error (a loss function)
reaches a (local) minimum. 
Taking into account that  
the environment variables have an inherent  noisy component 
and the organism receptors sense these variables with some error
one has to deduce that the training data sets are always noisy ($a_0 > 0$).
The same time it is desirable to be able to predict 
a trend of the environment changes. Thus, we again encounter
the same problem of predicting smooth dynamics based on 
the input sequences with nonzero noise amplitude,
and the problem is solved due to unexpected "predictive 
smoothing" feature of trained PRNs.

\section{Discussion}

A traditional approach
of predicting neural networks holds that a trained network
is expected to predict time series with the same
qualitative features as those characterizing the 
input data.
This view can be justified when the input signal to the network is smooth,
{\it i.e.}, when the input signal has the same noise level
as the actual trajectory and the network is trained on such signals $a_0=a_i=0$. 
When the input signal is noisy and the network trained
on the smooth signal ($a_i > a_0=0$), it fails to predict with reasonable 
accuracy. 

Thus the paradigm should be changed to one that allows both
the training data sets and input signals have nonzero noise component
$a_0, \; a_i > 0$ reflecting real case scenario.
The numerical experiments presented above demonstrate that 
PRNs trained with noisy data sets $(a_0 > 0)$
predict smooth trajectories for both smooth $(a_i = 0)$ and perturbed $(a_i > 0)$ inputs.
When the number of neurons is large enough the predicted trajectories 
appear to be quite close to the actual trajectories $\bm f(t)$.
We argue that prediction of smooth trajectories is  
mainly due to the training procedures but it also can be 
related to predictive algorithm choice.


The application of both MW and EW predictive algorithms as well as their memoryless
counterparts
for the trained PRN with sufficient number of neurons would produce a
sequence of points that are positioned close to the actual trajectory $\bm f(t)$.
The statistical arguments used in the manuscript allow to explain "smoothness" of the 
predicted dynamics for the well trained networks. It might happen that 
this is not the only reason for such a behavior and the actual explanation requires
much deeper insight into mathematical facet of the problem. 

This statement is based on the following observation -- 
it was reported in \cite{Rub2020a} that 
a {\it trained} PRN with a small number of neurons that cannot successfully predict
an actual trajectory still generates a {\it smooth} curve. 
The same time preliminary numerical experiments show that on the contrary {\it untrained} PRNs with  
randomly set trainable parameters demonstrate periodic and even chaotic
transient behavior as well as stable periodic long time predicted dynamics.
One can view that as a mathematical curiosity which is irrelevant  
from the biological point of view as a low prediction quality
diminishes survival probability that creates an evolutionary pressure 
forcing an organism to increase the size (number of neurons) of predictive network.

What is the difference between the trained and untrained network?
From the mathematical perspective PRNs dynamics
is governed by a discrete time map (\ref{ML_s}) with the corresponding 
set of parameters.
Random parameters of the untrained network in a process of
training are replaced by another set for the trained network.
As we observe the training "forces" a network to generate the first
point of the predicted trajectory to be in a small vicinity of 
the unperturbed trajectory. On the contrary, the corresponding point produced by
an untrained PRN lies much further off the actual curve.
The consecutive steps of prediction by EW algorithm depend on the 
inner dynamical properties of the map (\ref{ML_s}).

This fact illuminates two-sided effect of a training procedure.
First, its main 
goal is to produce 
a proper location of a single predicted point in "seq-to-one" paradigm.
Additionally the same training process gives rise to 
a specific discrete time map (\ref{ML_s}) that determines 
recursively predicted dynamics beyond the first point. 
Number of qualitatively different predicted trajectories depends 
only on the nonlinear map  (\ref{ML_s}) while the selection between them 
is due to position of the first predicted point (mostly determined
by the input sequence). In the framework of dynamical system theory
different trajectories correspond to different attractors of (\ref{ML_s})
each one having its unique basin of attraction. The number of attractors and
complexity of the basin topology usually 
grows with increase in number of parameters (network size).

When the number of neurons is small the trajectories are very simple, e.g.,
smooth approach to a steady state point and the network fails to reproduce a desired dynamics.
If even such unsuccessful training leads 
to generation of some smooth
dynamics by (\ref{ML_s}) then this property might be a main 
reason for smooth prediction made by well trained PRNs 
and it deserves separate investigation.

In this manuscript we consider how perturbations of
input sequences affect robustness of predictions made by
PRNs trained with data sets of various noise level.
In other words, we investigate how an external
perturbation influences the network performance. 
It would be instructive to consider results of 
trained network internal perturbation effected 
by disabling of one or more neurons. 
From the neuroscience perspective such a scenario 
is quite possible in natural neural networks. 
In more relevant problem formulation when a network consists of several recurrent layers
this question becomes more complex as one might expect that damage
of a few neurons in the inner layers would have less effect on the prediction 
quality compared to switching the neurons off in the 
first or last layer. Investigation 
of multilayer predictive network dynamics will be published elsewhere.



\vskip 0.6cm
\noindent
{\Large \bf Appendix}

\appendix

\section*{Memoryless vs. EW/MW prediction algorithm}

\setcounter{equation}{0}
\renewcommand{\theequation}{A\arabic{equation}}

The first prediction round of the algorithm is the same as in
MW/EW algorithms -- 
a sequence $\bm X^{1}$ of length $m$ is fed
into the network that produces $\bm s_{m}^1$ 
to generate the prediction $\bar{\bm x}_{m+1} = \bm L(\bm s_m^1)$.

In regular EW algorithm for the second round one forms a new expanded sequence $\bm X^{2}$ of length $m$
$\bm X^{2}= \{\bm x_1,\bm x_2,\ldots,\bm x_m, \bar{\bm x}_{m+1}\}$ and feeds it into
the network to produce 
\be
\bm s_{m+1}^2 = \bm F({\bm x}_{m+1}^2, \bm s_{m}^2)
= \bm F(\bar{\bm x}_{m+1}, \bm s_{m}^1).
\label{MLr2}
\ee
Here we use the facts that ${\bm x}_{m+1}^2 = \bar{\bm x}_{m+1}$ and 
$\bm s_{m}^2 = \bm s_{m}^1$ as the first $m$ elements in $\bm X^{1}$ and
$\bm X^{2}$ coincide. We observe that both arguments 
in $\bm F(\bar{\bm x}_{m+1}, \bm s_{m}^1)$ are found 
at the first prediction round.
Thus we can retain the inner state $\bm s_m^1$ as the current state 
of the network and use the 
prediction $\bar{\bm x}_{m+1}$ as an input to RNN to 
compute  $\bm s_{m+1}^2$ and to arrive at $\bar{\bm x}_{m+2} = \bm L(\bm s_{m+1}^2)$.

In the third round we have
$\bm X^{3}= \{\bm x_1,\bm x_2,\ldots,\bm x_m,\bar{\bm x}_{m+1}, \bar{\bm x}_{m+2}\}$
and obtain
\be 
\bm s_{m+2}^3 = \bm F(\bm x_{m+2}^3, \bm s_{m+1}^3)=
\bm F(\bar{\bm x}_{m+2}, \bm s_{m+1}^2)=
\bm F(\bm L(\bm s_{m+1}^{2}),\bm s_{m+1}^{2}) ,
\quad 
\bar{\bm x}_{m+3} = \bm L(\bm s_{m+2}^3).
\label{MLr3}
\ee
Continue the same procedure into $k$-th prediction round tp find
\be
\bm s_{m+k-1}^k = \bm F(\bm x_{m+k-1}^k, \bm s_{m+k-2}^{k})
= \bm F(\bar{\bm x}_{m+k-1}, \bm s_{m+k-2}^{k-1}) = 
\bm F(\bm L(\bm s_{m+k-2}^{k-1}),\bm s_{m+k-2}^{k-1}) ,
\quad
\label{ML}
\ee
We observe that the results generated by the EW approach can 
be obtained with its ML version that does not require repeated 
construction of the input sequences of increasing length.

Note that the last element of the 
inner state sequence at the round $k$ reads $\bm s_{m+k-1}^k$ 
and the superscript $k$ might be dropped.
Rewrite (\ref{ML}) as
\be
\bm s_{m+k-1} = \bm F(\bar{\bm x}_{m+k-1}, \bm s_{m+k-2})
= \bm F(\bm L(\bm s_{m+k-2}), \bm s_{m+k-2}) = 
\bm H(\bm s_{m+k-2}),
\label{ML_s}
\ee
that represents a vector discrete map transforming 
the last state $\bm s_{m+k-2}$ of the 
input sequence $\bm X^{k-1}$ into a similar term
$\bm s_{m+k-1}$ in the next predictive round.
As soon as the first prediction round is over 
and the value $\bm s_{m}$ is computed the predicted sequence can be obtained
as the result of recurrent application of the map (\ref{ML_s}) to 
$\bm s_{m}$. In other words, the prediction is determined by the 
{\it autonomous} dynamics of the network.
It should be underlined here that if the network is untrained 
or when the number of neurons is too small to make 
a reliable prediction the map (\ref{ML}) will generate
the same sequence due to EW approach.

Considering the "moving window" algorithm first note that 
if the network it well trained, {\it i.e.},
the predicted value is close to the actual trajectory then
EW version is preferable over MW approach as it uses the 
larger input sequences. In such a case one expects that 
the predictions of these two algorithms should not differ
significantly and thus the ML algorithm can be used too.
Moreover, the numerical experiments performed
with untrained and small size networks that do not
demonstrate high predictive quality show that
both MW and EW (ML) algorithms still generate 
 predicted sequences that are very close to each other.


\end{document}